\PassOptionsToPackage{export}{adjustbox}
\documentclass{article}

\PassOptionsToPackage{numbers, compress}{natbib}
\usepackage[final, main]{neurips_2026}
\usepackage[T1]{fontenc}
\input{macros}

\title{Cooperative Multi-Agent Reinforcement Learning via Epigraph-Form Guided Exploration}

\author{%
  Sungil Son\textsuperscript{1, *}
  \quad Hoseong Jung\textsuperscript{1, *}
  \quad Dahyun Oh\textsuperscript{1, *}
  \quad H. Jin Kim\textsuperscript{1, $\dagger$}\\
  \textsuperscript{1}
  Seoul National University \\
  \texttt{\{aldridge99, ghtjdaleka, qlass33, hjinkim\}snu.ac.kr} \\
}

\let\oldaddcontentsline\addcontentsline
\renewcommand{\addcontentsline}[3]{}

\begin{document}

\maketitle

\begingroup
\renewcommand{\thefootnote}{}
\footnotetext{%
  \textsuperscript{*}Equal contribution.
  \textsuperscript{$\dagger$}Corresponding author.
}
\endgroup

% ================================================================
\vspace{-5pt}
\begin{abstract}
Discovering cooperative behavior in multi-agent reinforcement learning (MARL) is challenging due to the combinatorial complexity of joint state-action spaces, which hinders the emergence of coordinated behaviors from trial-and-error alone.
Intrinsic rewards are often used to aid discovery, but naively combining them with team objectives can distort the learning signal, compromising task performance.
In~this paper, we propose \textsc{EFXPLORER}, a constrained exploration framework that maximizes exploration objectives subject to a constraint that preserves established task performance.
We solve the resulting problem using an epigraph reformulation that introduces adaptive exploration budgets.
This approach separates intrinsic rewards from task objectives and regulates exploration through task feasibility.
To~further encourage diverse and temporally extended exploration, we incorporate a successor distance-based intrinsic reward that captures long-horizon dependencies.
Empirically, our method outperforms strong baselines and induces novel cooperative strategies across SMAX, VMAS, and MPE benchmark suites.

Our code and demos are available at 
\texttt{\href{http://efxplorer-marl.github.io/}{\textcolor{purple}{http://efxplorer-marl.github.io/}}}
\end{abstract}
% ================================================================

\section{Introduction}\label{sec:intro}
Cooperative Multi-Agent Reinforcement Learning (MARL) aims to enable a team of agents to discover and execute joint behaviors that achieve a shared task objective~\cite{albrecht2024multi}.
However, in many team-based scenarios, successful coordination occupies only a tiny subspace of the joint state-action space.
Coordination is relational and context-dependent, so it is often ambiguous to design an appropriate task reward.
Consequently, cooperative MARL increasingly leverages task-agnostic exploration signals to uncover coordination~\cite{gronauer2022multi, du2023review, hao2023exploration}.

\begin{figure}[h!]
\includegraphics[width=\textwidth]{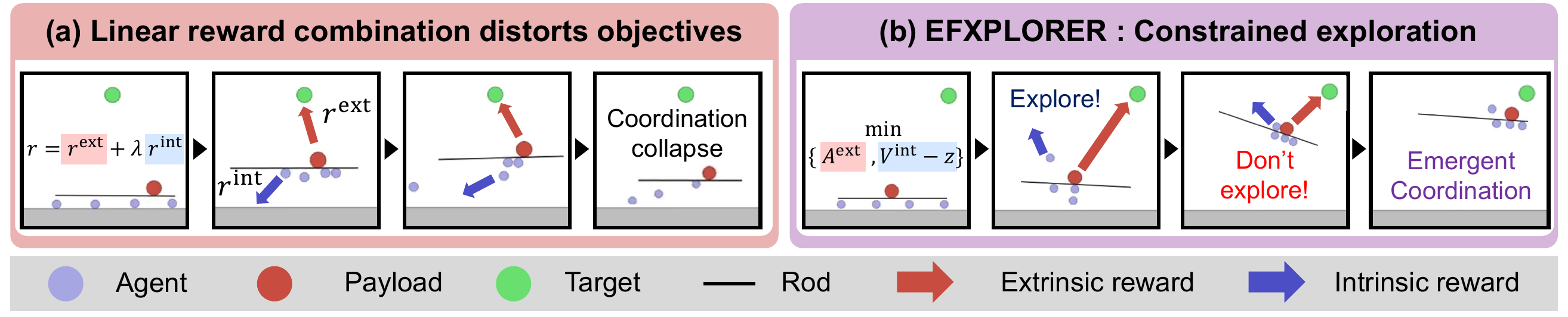}
\caption{Cooperative balancing task with four agents transporting a payload on a rod to a target.
(a) Naive linear reward combinations induce objective misalignment, destabilizing coordination. (b) \textsc{EFXPLORER} leverages a min-function to conservatively gate exploration, enabling coordination to emerge progressively without sacrificing task performance.
}
\label{fig:main-figure}
\vspace{-15pt}
\end{figure}

A widely used remedy is to add intrinsic exploration bonuses to compensate for ambiguous task feedback~\cite{li2023two}.
However, this often distorts the learning signal, as exploration incentives can overshadow task objectives (Fig.~\ref{fig:main-figure}).
Furthermore, an agent's greedy exploration triggers a shift toward low-reward regions, which creates a non-stationary environment for teammates~\cite{panait2005cooperative, yang2025game}.
Partners often mistake this temporary drop for failure and prioritize individual stability, causing the joint coordination to collapse.
This motivates a conservative exploration framework that preserves cooperative behaviors.

In this paper, we propose \textsc{EFXPLORER}, a conservative exploration framework for cooperative MARL.
Our core objective is to maximize intrinsic exploration of each agent while constraining the joint policy to preserve established task performance.
We address this optimization via an epigraph reformulation~\cite{boyd2004convex}, which introduces an auxiliary variable serving as attainable exploration budgets.
Unlike a primal--dual relaxation, our epigraph-based formulation avoids learning a tuning-sensitive dual multiplier whose magnitude rescales the policy gradient.
Instead, intrinsic rewards remain separate from the scalarized task return, so exploration is gated by feasibility rather than directly biasing task-oriented policy updates.

We solve this reformulated objective via an alternating maximization scheme.
The central technical step is to convert the episode-level performance non-degradation constraint into a stepwise min-bottleneck between task progress and intrinsic surplus.
Given fixed exploration budgets, the inner loop updates the policy using the conservative epigraph objective, suppressing intrinsic updates when the task-improvement signal is insufficient.
The outer loop then expands the budget based on recent high-novelty rollouts, contingent on preserved task performance.
\textsc{EFXPLORER} therefore expands the attainable novelty frontier progressively, rather than pursuing exploration through unconstrained reward maximization.

To drive effective exploration, we design an intrinsic objective based on successor distance~\cite{myers2024learning}.
Unlike similarity metrics that are sensitive to local noise, this measure captures the long-horizon connectivity of the state space, driving agents toward temporally distinct states~\cite{jiang2025episodic}.
We define successor distance on individual state spaces to capture scalable, agent-centric novelty without joint-space complexity.
Since such local exploration inherently risks destabilizing coordination, \textsc{EFXPLORER} is essential to strictly regulate these incentives.

\textbf{Our contributions include:}
1) We propose \textsc{EFXPLORER}, a conservative exploration framework that reformulates cooperative exploration as an epigraph-based constrained optimization problem and derives a stepwise bottleneck for feasibility-preserving policy updates;
2) We instantiate the exploration objective using a scalable successor distance defined on individual state spaces, capturing long-horizon dependencies without joint-space complexity;
3) We empirically validate EFXPLORER on SMAX~\cite{rutherford2024jaxmarl}, VMAS~\cite{bettini2022vmas}, and MPE~\cite{lowe2017multi} benchmarks, demonstrating consistent improvements over state-of-the-art baselines and applicability across diverse cooperative tasks and agent configurations.
% ================================================================
\section{Background}\label{sec:background}
\subsection{Cooperative Multi-Agent Reinforcement Learning}\label{subsec:2.1}
We formulate the cooperative multi-agent task as a Decentralized Partially Observable Markov Decision Process (Dec-POMDP) \cite{oliehoek2016concise} $\langle \mathcal{N}, \mathcal{S}, \Omega, \mathcal{O}, \mathcal{A}, \mathcal{P}, \mathcal{R}, \gamma \rangle$, 
where $\mathcal{N} = \{1,\dots,n \}$ is the set of $n$ agents. 
$s_t \in \mathcal{S}$ represents the global state of the environment at timestep $t$.
We define $s_t=[s_t^1, \dots, s_t^n, s_t^{\text{env}}]$, where $s_t^i\in\mathcal{S}_i$ is the individual state of the agent $i$ and $s_t^{\text{env}}$ denotes the environment-specific state.
At each timestep $t\geq 0$, the agent $i$ receives a local observation $o_t^i \in \Omega$ via the observation function $\mathcal{O}(s_t, i)$ and selects an action $a_t^i\in\mathcal{A}$.
The joint action $\mathbf{a}_t=(a_t^1, \dots, a_t^n)\in \mathcal{A}\equiv\prod\mathcal{A}_i$ triggers a transition to the next state $s_{t+1}$ with transition probability $\mathcal{P}(s_{t+1}|s_t, \mathbf{a}_t)$.
The agents' collaborative behavior is defined by a joint policy $\boldsymbol{\pi}=(\pi^1, \dots, \pi^n)$, where $\pi^i(a^i|o^i)$ is the individual policy for agent $i$.

All agents aim to maximize a shared extrinsic team reward $r_t^\text{ext}=\mathcal{R}(s_t, \mathbf{a}_t)$.
To encourage exploration, each agent~$i$ additionally receives an intrinsic reward $r_{i,t}^{\text{int}}$ measuring state novelty or reachability~\cite{bellemare2016unifying,jiang2025episodic}.
We denote the value functions corresponding to each reward as $V^{\text{ext}}_{t}$ and $V^{\text{int}}_{i,t}$ for agent $i$ at timestep $t$.
Accordingly, the extrinsic and intrinsic objectives under the joint policy $\boldsymbol{\pi}$ are defined as:
\begin{equation}
    J_\text{ext}(\boldsymbol{\pi})=\mathbb{E}_{\boldsymbol{\pi}}[\sum_{t=0}^{\infty}\gamma^tr_t^{\text{ext}}], \;
    J_\text{int}(\boldsymbol{\pi})=\sum_{i=1}^n\mathbb{E}_{\pi^i}[\sum_{t=0}^{\infty}\gamma^tr_{i,t}^{\text{int}}].
    \label{eq:ext_and_int_objective}
\end{equation}
A conventional approach~\cite{jaques2019social, zheng2021episodic, li2023two} balances these objectives by maximizing a weighted sum $    J_\text{conv}(\boldsymbol{\pi})=J_\text{ext}(\boldsymbol{\pi})+\beta J_\text{int}(\boldsymbol{\pi})$, where $\beta\geq 0$ controls the exploration-exploitation trade-off.
However, this linear combination distorts the extrinsic objective, potentially hindering coordination~\cite{chen2022redeeming}.
To~address this, we propose a constrained optimization framework that maximizes exploration while preserving extrinsic task performance.
\subsection{Epigraph Form for Constrained Optimization}\label{subsec:2.2}
The epigraph form~\cite{boyd2004convex} is a standard reformulation technique that introduces an auxiliary scalar variable to express a constrained optimization problem via a max--min structure.\footnote{While the standard epigraph form is defined for minimization~\cite{boyd2004convex}, we adopt the maximization (hypograph) formulation and adjust sign conventions accordingly.}
Consider a general constrained optimization problem:
\begin{equation}
    \max_x J(x) \quad \text{s.t.} \quad h(x)\geq 0.
    \label{eq:standard_optim}
\end{equation}
The epigraph form introduces an auxiliary variable $z\in\mathbb{R}$ to represent the lower bound of the objective $\max_{x,z} z \; \text{s.t.} \; h(x) \geq 0, J(x) \geq z$.
This can be transformed into an equivalent hierarchical optimization problem:
\begin{equation}
    \max_{z} z \quad \text{s.t.} \quad \max_x\min\{\eqnmarkbox[red]{feasibility}{h(x)}, \eqnmarkbox[vibrant_blue]{objective}{J(x)-z}\}\geq 0.
\end{equation}
This decomposition yields an unconstrained \textit{inner problem} over $x$ for a fixed $z$ and an \textit{outer problem} that adjusts the scalar variable $z$.
A key benefit of this form is that the \textit{\textcolor{red}{original feasibility constraint}} and the \textit{\textcolor{blue}{objective lower-bound constraint}} appear in a unified way through the $\min$ operator.
As a result, it avoids introducing a sensitive scalar weight to trade off feasibility and optimality.

Epigraph-based reformulations have recently been used in constrained RL to enforce safety or cost requirements while avoiding manually tuned trade-off coefficients~\cite{so2023solving, zhang2025solving}.
Unlike safety-oriented settings, where constraint violations are defined at individual timesteps, our focus is return-level non-degradation evaluated over episodic cooperative performance.
This motivates our use of the induced max--min structure to couple task improvement with exploration surplus (Section~\ref{subsec:3.1}).
\subsection{Successor Distance for Long-Horizon Exploration}\label{subsec:2.3}
Prevailing intrinsic rewards are driven by local prediction~\cite{pathak2017curiosity} or representation errors~\cite{burda2018exploration}, which often emphasize short-term novelty.
In contrast, temporal distance captures underlying long-horizon structures.
Following~\cite{myers2024learning}, we define the discounted occupancy measure~$p_\gamma^{\boldsymbol{\pi}}(y|x)=(1-\gamma)\sum_{k=0}^\infty \gamma^k\, p^{\boldsymbol{\pi}}(s_k=y|s_0=x)$, and formulate the Successor Distance (SD) as:
\begin{equation}
    d^{\boldsymbol{\pi}}_{\text{SD}}(x, y) = \log \!\left( \frac{p^{\boldsymbol{\pi}}_\gamma(s_f=y \mid s_0= y)}{p^{\boldsymbol{\pi}}_\gamma(s_f=y \mid s_0=x)} \right).
    \label{eq:successor_distance}
\end{equation}

SD serves a quasimetric notion of temporal proximity~\cite{myers2024learning, myers2025offline} and has been used to define intrinsic rewards for exploration~\cite{jiang2025episodic}.
In cooperative MARL, directly computing SD-based novelty over the joint state space suffers from the curse of dimensionality~\cite{zamboni2025towards}.
This motivates the adoption of factorized agent-centric SD estimates to compute intrinsic rewards within our framework (Section~\ref{subsec:3.2}).

% ================================================================
\section{Methodology}\label{sec:method}
\begin{figure}[t!]
\includegraphics[width=\textwidth]{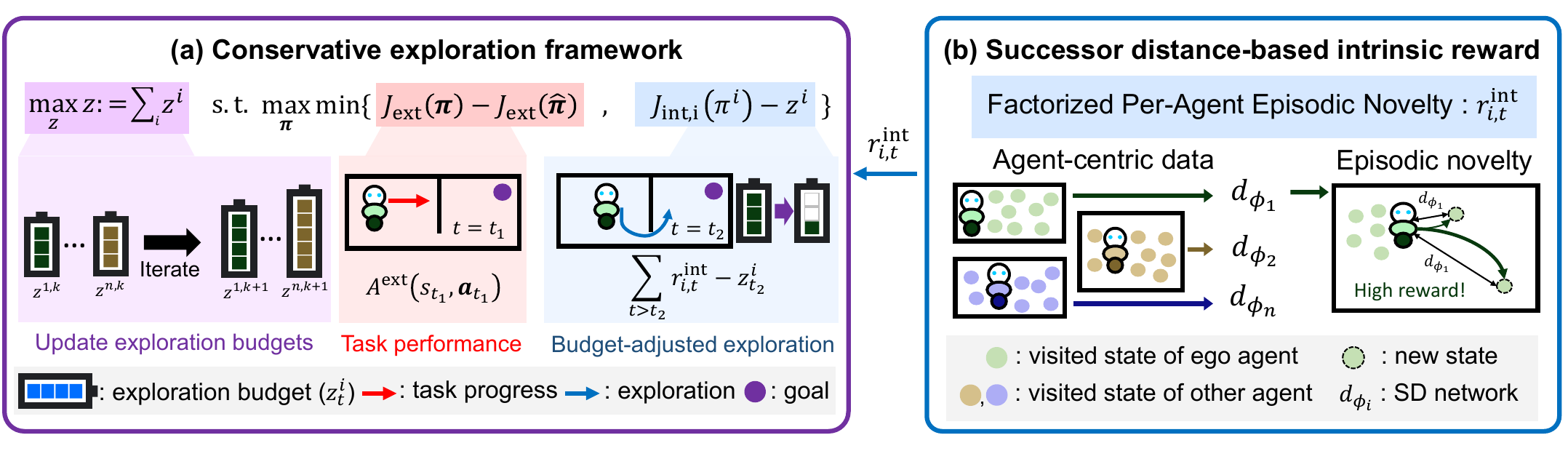}
\caption{Overview of \textsc{EFXPLORER}. (a) Conservative exploration (Sec.~\ref{subsec:3.1}): The epigraph objective uses a $\min$ operator to balance task progress and budget-adjusted exploration while an outer loop adapts the budget $z$.
(b) Factorized episodic novelty (Sec.~\ref{subsec:3.2}): Per-agent Successor Distance (SD) networks estimate temporal proximity to reward individual novelty ($r^{\text{int}}_{i,t}$).
\vspace{-15pt}
}
\label{fig:efxplorer}
\end{figure}

This section introduces \textsc{EFXPLORER}, a conservative exploration framework for MARL based on an epigraph reformulation, as shown in Fig.~\ref{fig:efxplorer}.
Our approach consists of three main components:
(i) an \textbf{epigraph-based formulation} that separates task and exploration signals and couples them conservatively (Sec.~\ref{subsec:3.1});
(ii) a \textbf{successor distance-based intrinsic reward} that captures long-horizon, temporally structured novelty (Sec.~\ref{subsec:3.2}); 
and (iii) a \textbf{practical multi-agent actor–critic implementation} with low-variance advantage estimation (Sec.~\ref{subsec:3.3}).
\subsection{Conservative Exploration via Epigraph Form}\label{subsec:3.1}
\paragraph{Epigraph Learning Framework.}
Our primary goal is to induce diverse exploration behaviors without compromising the established cooperative task performance.
We formulate this as a constrained optimization problem where the intrinsic return is maximized subject to the constraint that extrinsic performance maintains the level of a reference joint policy $\hat{\boldsymbol{\pi}}$ (e.g., the previous iterate):

\begin{equation}
\max_{\boldsymbol{\pi}\in\Pi} \sum_{i\in\mathcal{N}} J_{i,\text{int}}(\pi^i)\quad \text{s.t.} \quad J_\text{ext}(\boldsymbol{\pi}) \geq J_\text{ext}(\hat{\boldsymbol{\pi}}),
\label{eq:constrained_optimization}
\end{equation}
where $J_\text{ext}$ is team-coupled and depends on the joint policy, while $J_{i, \text{int}}$ is agent-specific to promote diverse exploration.
We assume synchronized joint updates: all factorized policies are updated in parallel from the same reference policy $\hat{\boldsymbol{\pi}}$, so the agent-wise constraints share one candidate joint policy rather than separate per-agent maximizers.
We justify this assumption in Appendix~\ref{sec:appendix_A}.

Using the epigraph reformulation introduced in Section~\ref{subsec:2.2}, we introduce an auxiliary variable $z^i \in \mathbb{R}$ for each agent.
Under the synchronized joint-update assumption, Eq.~\eqref{eq:constrained_optimization} can be reformulated as a joint maximization problem:

\vspace{-0.1in}
\begin{equation}
    \max_{z} z:= \sum_{i\in\mathcal{N}} z^i \quad
    \text{s.t.}\quad  \max_{\boldsymbol{\pi} \in \Pi}\min\Big\{J_\text{ext}(\boldsymbol{\pi})-J_\text{ext}(\hat{\boldsymbol{\pi}}), J_{i,\text{int}}(\pi^i)-z^i\Big\}\ge 0, \quad 
    \forall i \in \mathcal{N}.
    \label{eq:epigraph_form}
\end{equation}

We interpret $z^i$ as a dynamic \textit{exploration budget} for agent~$i$.
Subtracting $z^i$ within the $\min$ operator conditions intrinsic updates on task feasibility, discouraging exploration when constraints are violated.
% Here, $z=\sum_{i\in\mathcal{N}} z^i$ aggregates agent-wise exploration budgets into a single team-level objective.
% This reformulation replaces manual coefficient tuning with a unified epigraph bottleneck, coupling exploration to task feasibility via the $\min$ operator.
This reformulation removes the explicit coefficient used to mix extrinsic and intrinsic objectives.
Exploration is instead coupled to task feasibility through the $\min$ operator.

To solve Eq.~\eqref{eq:epigraph_form}, we employ an alternating maximization scheme.
Since the outer objective is to maximize $z$, we alternate between updating $\boldsymbol{\pi}$ to expand the feasibility margin (\textit{inner loop}) and increasing the exploration budget $z$ while maintaining feasibility (\textit{outer loop}).

\paragraph{Inner Optimization.}
For a fixed set of auxiliary variables $\{z^i\}$, the inner problem in Eq.~\eqref{eq:epigraph_form} updates each agent's policy $\pi^i$ to expand the feasibility margin:
\begin{equation}
    \min\Big\{ \eqnmarkbox[red]{ext}{J_{\text{ext}}(\boldsymbol{\pi})-J_{\text{ext}}(\hat{\boldsymbol{\pi}})},\; \eqnmarkbox[vibrant_blue]{int}{J_{i,\text{int}}(\pi^i)-z^i} \Big\}.
    \label{eq:inner_optim}
\end{equation}
We first handle the left-hand term, which measures \textit{\textcolor{red}{task progress}} relative to the reference joint policy~$\hat{\boldsymbol{\pi}}$.
To do so, we use the policy improvement indicator~\cite{kakade2002approximately, schulman2015trust} based on the advantage function $A_{\hat{\boldsymbol{\pi}}}^{\text{ext}}(s,\mathbf{a})=Q_{\hat{\boldsymbol{\pi}}}^{\text{ext}}(s,\mathbf{a})-V_{\hat{\boldsymbol{\pi}}}^{\text{ext}}(s)$.
Denoting the performance difference as $\Delta J_\text{ext}(\boldsymbol{\pi})\triangleq J_\text{ext}(\boldsymbol{\pi})-J_\text{ext}(\hat{\boldsymbol{\pi}})$, we apply the performance difference identity to approximate this term via trajectories from~$\boldsymbol{\pi}$:

\vspace{-0.1in}
\begin{equation} 
    \Delta J_\text{ext}(\boldsymbol{\pi})
    \approx \mathbb{E}_{\tau \sim \boldsymbol{\pi}} \!\left[\sum_{t\ge0} \gamma^t A_{\hat{\boldsymbol{\pi}}}^\text{ext}(s_t, \mathbf{a}_t)\right].
    \label{eq:performance_difference}
\end{equation}

Based on this identity, a conservative sufficient condition for non-degradation is that the policy support contains only joint actions with non-negative reference advantage:
\begin{equation}
    A_{\hat{\boldsymbol{\pi}}}^\text{ext}(s,\mathbf{a}) \ge 0, \quad \forall s,\ \forall \mathbf{a}\in\mathrm{supp}(\boldsymbol{\pi}(\cdot|s)).
    \label{eq:stepwise_perform_identity}
\end{equation}

Under this condition, we provably have $J_\text{ext}(\boldsymbol{\pi})\geq J_\text{ext}(\hat{\boldsymbol{\pi}})$ (see Theorem~\ref{thm:suff_stepwise} in Appendix~\ref{sec:appendix_A}).
Since strictly enforcing this condition over the entire state-action space is intractable, we approximate it by optimizing the stepwise surrogate over sampled trajectories~\cite{kakade2002approximately}.

We next consider the right-hand term $J_{i,\text{int}}(\pi^i)-z^i$, which represents agent $i$'s \textit{\textcolor{blue}{intrinsic surplus}} relative to its exploration budget.
To obtain a stepwise surrogate analogous to the reference extrinsic advantage, we incorporate the remaining exploration budget into the state space.
Specifically, we introduce the discounted budget dynamics:
\begin{equation}
    \gamma z_{t+1}^i=z_{t}^i-r^{\text{int}}_{i,t},\qquad z^i_0=z^i,
\label{eq:budget_dynamics}
\end{equation}
so that the augmented state $\tilde{s}^i_t=(s_t,z^i_t)$ has the Markov property~\cite{carrara2019budgeted, sootla2022saute}.
With this augmentation, we can express the intrinsic surplus-to-go $S_i^\text{int}$ in a stepwise formulation:
\begin{equation}
    S_i^{\text{int}}(\tilde{s}_t^i,\mathbf{a}_t)
    \;\triangleq\;
    \mathbb{E}_{\tau\sim\boldsymbol{\pi}}
    \Big[\sum_{k=t}^{\infty}\gamma^{k-t} r_{i,k}^{\text{int}} - z_t^i \ | \tilde{s}_t^i,\,\mathbf{a}_t\Big].
    \label{eq:surplus_def}
\end{equation}
We unify the stepwise task improvement $A^\text{ext}_{\hat{\boldsymbol{\pi}}}$ and intrinsic surplus $S_i^\text{int}$ into the epigraph bottleneck:
\begin{equation}
    \mathcal{B}_{i}^{\text{epi}}(\tilde{s}_t^i,\mathbf{a}_t)
    \triangleq
    \min\!\Big\{A_{\hat{\boldsymbol{\pi}}}^{\text{ext}}(s_t,\mathbf{a}_t),\; S_{i}^{\text{int}}(\tilde{s}_t^i,\mathbf{a}_t)\Big\}.
    \label{eq:epigraph_bottleneck}
\end{equation}

\begin{proposition}[Conservative Epigraph Feasibility]\label{prop:epi_surr}
Enforcing the condition $\mathcal{B}_i^\text{epi}(\tilde{s}_t^i,\mathbf{a}_t)\ge0$ for all visited states and actions provides a conservative sufficient condition for feasibility of the original epigraph constraints.
\end{proposition}

Guided by this guarantee, we define the global bottleneck objective: $J_i^\text{epi}(\boldsymbol{\pi}) \triangleq \mathbb{E}_{\tau}[\sum_{t} \gamma^t \mathcal{B}_{i}^{\text{epi}}(\tilde{s}_t^i,\mathbf{a}_t)]$.
Directly differentiating it is intractable due to evolving distribution shifts~\cite{schulman2015trust, schulman2017proximal}.
Thus, we optimize a tractable local surrogate by fixing both $A_{\hat{\boldsymbol{\pi}}}^{\text{ext}}$ and $S_i^\text{int}$ evaluated under the reference policy $\hat{\boldsymbol{\pi}}$~\cite{yu2022surprising}.
This yields a local surrogate of the feasibility margins:
\begin{equation}
    L_i^\text{epi}(\theta)
    =
    \hat{\mathbb{E}}_t
    \left[
    \rho^i_t(\theta)\,
    \mathcal{B}_{i}^{\text{epi}}(\tilde{s}_t^i,\mathbf{a}_t)
    \right],
    \quad
    \rho^i_t(\theta)=
    \frac{\pi^i_\theta(a^i_t|o^i_t)}
    {\hat{\pi}^i(a^i_t|o^i_t)} .
    \label{eq:local_epi_surrogate}
\end{equation}

\begin{theorem}[Local Epigraph Policy Gradient]\label{thm:epi_pg}
Let $\pi_\theta^i$ be the parameterized policy of agent $i$, and let $\hat{\boldsymbol{\pi}}$ be the fixed reference policy.
The gradient of the local epigraph surrogate $L_i^\text{epi}(\theta)$ is
\begin{equation}
    \nabla_\theta L_i^{\mathrm{epi}}(\theta) = \hat{\mathbb{E}}_t
    \left[\rho^i_t(\theta)\nabla_\theta \log \pi^i_\theta(a^i_t|o^i_t)A^{\mathrm{epi},\hat{\boldsymbol{\pi}}}_{i,t}\right].
    \label{eq:local_epi_pg}
\end{equation}
where $A^{\text{epi},\hat{\boldsymbol{\pi}}}_{i,t}=\mathcal{B}_{i}^{\text{epi}}(\tilde{s}_t^i,\mathbf{a}_t)-b_i(\tilde{s}_t^i)$ for any action-independent baseline $b_i$.
\end{theorem}

Theorem~\ref{thm:epi_pg} shows that the fixed-reference epigraph update reduces to a standard likelihood-ratio gradient weighted by an epigraph advantage.
In Section~\ref{subsec:3.3}, we construct a GAE-style estimator~\cite{schulman2015high} for low-variance learning of $A^{\mathrm{epi},\hat{\boldsymbol{\pi}}}$, and use it as the advantage signal in PPO-style actor updates~\cite{schulman2017proximal}.

\paragraph{Outer Optimization.}
The outer loop maximizes the team exploration budget $z=\sum_{i\in\mathcal{N}} z^i$ while preserving inner feasibility.
At iteration $k$, we collect rollouts from $M$ parallel environments.
For each rollout instance $m \in \{1, \dots, M\}$, the candidate budget is the empirical intrinsic return: $\hat{z}_{0}^{i,k,m} \; \triangleq\; \hat{G}_{i,k,m}^{\text{int}}\;=\sum_{t=0}^{T_m-1}\gamma^t r^{\text{int}}_{i,t}$, representing the tightest exploration bound satisfied empirically.

We subsequently evaluate inner feasibility after the policy update.
To guard the outer update against approximation errors inherent in critic-based surrogates, we verify the extrinsic constraint in Eq.~\eqref{eq:constrained_optimization} using Monte Carlo task returns.
We define the feasible set $\mathcal{F}_k$ as the set of rollout indices that outperform the reference performance level:
\begin{equation}
    \mathcal{F}_k \; \triangleq\; \Big\{m: \hat{J}^\text{ext}_{k,m} > \bar{J}^\text{ext}_k\Big\}, \quad \hat{J}_{k,m}^\text{ext}\triangleq \sum_{t=0}^{T_m-1}\gamma^t r_t^\text{ext}.
    \label{eq:outer_feasible_set}
\end{equation}

Here, $\bar{J}^\text{ext}_k$ is a running reference (e.g., the moving average of the best observed returns).
Among these feasible candidates, we select the rollout that maximizes the collective exploration budget:
\begin{equation}
    m^\ast \;\in\; \argmax_{m\in\mathcal{F}_k} \ \sum_{i\in\mathcal{N}} \hat{z}_{0}^{i,k,m}.
    \label{eq:select_best_rollout}
\end{equation}
To reduce the variance and avoid outliers, we apply a soft update to the rollout-initial budgets $z_{0}^{i,k}$:
\begin{equation}
    z_{0}^{i,k+1} \leftarrow \alpha_z \cdot \hat{z}_{0}^{i,k,m^\ast} + (1-\alpha_z)\cdot z_{0}^{i,k} \quad \forall i \in \mathcal{N},
    \label{eq:init_z_ema}
\end{equation}
where $\alpha_z\in[0,1]$ controls the responsiveness of outer optimization.
If no feasible rollout exists (i.e., $\mathcal{F}_k=\emptyset$), we keep the previous budgets, $z_{0}^{i,k+1}\leftarrow z_{0}^{i,k}$.
\subsection{Successor Distance-based Intrinsic Reward}\label{subsec:3.2}
We instantiate the intrinsic objective $J_{i, \text{int}}$ using temporal distance novelty based on successor distance (we refer to Sec.~\ref{subsec:2.3} for the definition)~\cite{myers2024learning, myers2025offline}.

\paragraph{Agent-Centric Successor Distance via Contrastive Learning.}
To bypass the combinatorial complexity of the joint state space, we factorize SD onto individual state space $\mathcal{S}_i\times\mathcal{S}_i\rightarrow\mathbb{R}$~\cite{zamboni2025towards}.
Each agent $i$ trains a separate quasimetric network~$d_{\phi_i}$ using local trajectories to approximate the marginal SD $d_\text{SD}^{\boldsymbol{\pi}}$~\cite{nimonkar2025self}.
We optimize this network via a symmetric InfoNCE objective over buffer samples~\cite{oord2018cpc}, parameterizing the energy as $f_{\psi_i}(x,y) = c_{\eta_i}(y) - d_{\phi_i}(x,y)$~\cite{myers2024learning}. 
Theoretical guarantees for this factorized metric are provided in Appendix~\ref{subsec:Appendix_SD}.

\paragraph{Factorized Episodic Novelty.}
To encourage exploring distant states, we define the intrinsic reward as the minimum learned temporal distance to an agent's episodic memory $\{s_k^i\}_{k=0}^{t-1}$~\cite{jiang2025episodic}: 
\begin{equation}
    r^{\text{int}}_{i,t}(s_t^i) = \min_{k \in [0,t)} d_{\phi_i}(s_k^i, s_t^i).
    \label{eq:indiv_novelty}
\end{equation}
While this formulation promotes active discovery, such independent exploration can disrupt established coordination, highlighting the need for our constrained approach.
\subsection{Practical Training Procedure}\label{subsec:3.3}
\paragraph{Epigraph GAE.}
To estimate the epigraph advantage $A^{\text{epi},\boldsymbol{\hat{\pi}}}$ in Theorem~\ref{thm:epi_pg}, we decouple value estimation into extrinsic $V^\text{ext}(s_t)$ and intrinsic $V_i^\text{int}(s_t^i)$ critics.
For a horizon $k\geq1$, let $\mathcal{G}_{t,k}^\text{ext}$ and $\mathcal{G}_{i,t,k}^\text{int}$ denote the $k$-step bootstrapped returns.
We define the extrinsic advantage as $\hat{A}^\text{ext}_{t,k} = \mathcal{G}^\text{ext}_{t,k}-V^\text{ext}_t$ and the budget surplus as $\hat{S}^\text{int}_{i,t,k}=\mathcal{G}^\text{int}_{i,t,k}-z_{t}^i$. 
By subtracting the state-dependent baseline ($V_{i,t}^\text{int}-z_t^i$) from this surplus, we recover the standard intrinsic advantage $\hat{A}^\text{int}_{i,t,k} {=} \mathcal{G}^\text{int}_{i,t,k}{-}V^\text{int}_{i,t}$. 
We construct the $k$-step epigraph advantage by selecting the gradient of the binding term:
\begin{equation}
    \hat{A}^{\text{epi}}_{i,t,k} = 
    \begin{cases} 
        \hat{A}^\text{ext}_{t,k} & \text{if } \hat{A}^\text{ext}_{t,k} \leq \hat{S}^{\text{int}}_{i,t,k} \\
        \hat{A}^{\text{int}}_{i,t,k} & \text{otherwise.}
    \end{cases}
    \label{eq:kstep_epi_adv}
\end{equation}
Finally, we aggregate these estimates into a single target using a $\lambda_\text{GAE}$-weighted return in~GAE~\cite{schulman2015high}:
\begin{equation}
    A^\text{epi-GAE}_{i,t}=(1-\lambda_\text{GAE})\sum_{k=1}^{T-t} \lambda_\text{GAE}^{k-1}\cdot \hat{A}^\text{epi}_{i,t,k}.
    \label{eq:final_epi_adv}
\end{equation}

\paragraph{Overall Algorithm.}
Our framework supports both centralized and decentralized critic architectures. 
We integrate the epigraph update into two standard baselines: MAPPO~\cite{yu2022surprising} and IPPO~\cite{de2020independent}, termed EFXPLORER-MAPPO and EFXPLORER-IPPO, respectively.
The complete training procedure, including the inner policy update and outer budget adaptation, is detailed in Appendix~\ref{sec:appendix_E}.
% ================================================================
\section{Experiments}\label{sec:experiments}
\begin{figure}[t!]
    %\vskip -0.1in
    \begin{center}
        \centerline{\includegraphics[width=1.0\linewidth]{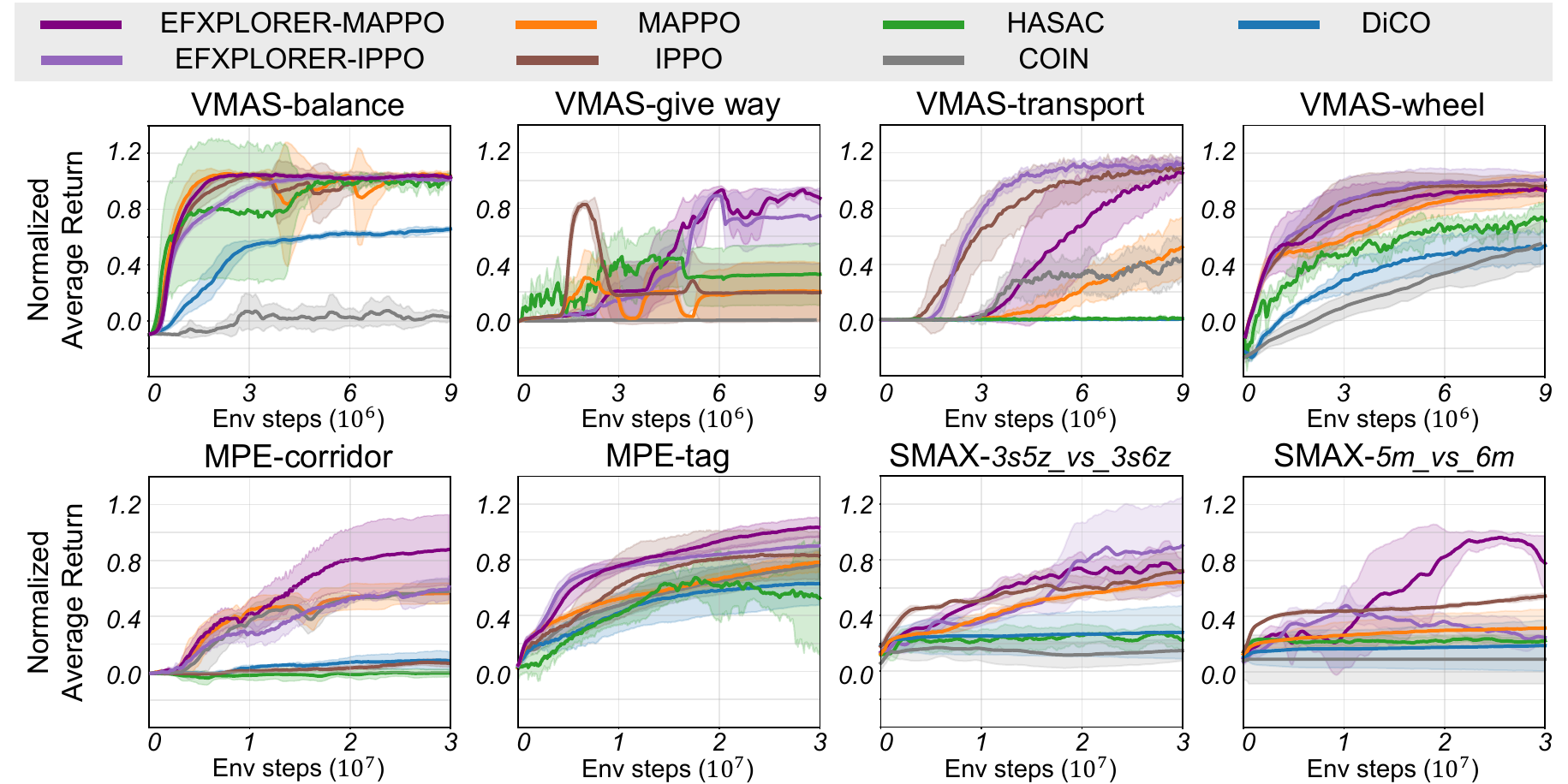}}
        \vskip -0.2in
    \end{center}
    \caption{Performance comparison of \textsc{EFXPLORER} with cooperative MARL baselines. 
    Solid lines show the mean over 5 random seeds, while shaded regions correspond to the standard deviation.}
    \label{fig:EFXPLORER_main_results}
    \vspace{-10pt}
\end{figure}

We evaluate \textsc{EFXPLORER} on cooperative multi-agent tasks from SMAX~\cite{rutherford2024jaxmarl}, VMAS~\cite{bettini2022vmas}, and MPE~\cite{lowe2017multi}.
For SMAX and MPE, we use the JaxMARL~\cite{rutherford2024jaxmarl} implementation and additionally define \textit{MPE-corridor}, a modified scenario designed to necessitate decentralized coordination in a narrow passage (see Fig.~\ref{fig:EFXPLORER_qual_analysis}).
For VMAS tasks, we employ the BenchMARL~\cite{bettini2024benchmarl} interface.
All methods use the same backbone architecture and a shared evaluation protocol under identical environment settings (MPE, SMAX: $3\times10^7$, VMAS: $9\times10^6$ environment steps).

Implementation details and hyperparameter specifications for \textsc{EFXPLORER} and all baselines are in Appendix~\ref{sec:appendix_E}.
We report normalized average team return throughout using task-wise linear normalization; task-specific baselines and environment details are provided in Appendix~\ref{sec:appendix_B}.

Our experiments aim to answer the following:
(1) Does \textsc{EFXPLORER} outperform cooperative MARL baselines? (Sec.~\ref{subsec:4.1})
(2) How much does each design component contribute to performance, and what role do the introduced design parameters play? (Sec.~\ref{subsec:4.2})
(3) Is the epigraph-based formulation superior to linear scalarization and Lagrangian alternatives? (Sec.~\ref{subsec:4.3})
(4) Do \textit{MPE-corridor} trajectories reveal adaptive trade-offs between exploration and task progress? (Sec.~\ref{subsec:4.4})\
\subsection{Comparison to Cooperative Baselines}\label{subsec:4.1}
We compare \textsc{EFXPLORER} against standard and exploration-oriented MARL baselines across centralized-training methods (e.g., \textbf{MAPPO}~\cite{yu2022surprising}, \textbf{COIN}~\cite{li2023two}, \textbf{HASAC}~\cite{liu2023maximum}) and independent learners (e.g., \textbf{IPPO}~\cite{de2020independent}, \textbf{DiCO}~\cite{bettini2024controlling}).
Fig.~\ref{fig:EFXPLORER_main_results} shows \textsc{EFXPLORER} improves normalized returns across centralized training and independent backbones.
Its epigraph bottleneck gates exploration, preserving task progress while inducing diverse behaviors.
Meanwhile, injecting intrinsic signals via social influence, entropy regularization, or enforced diversity (\textbf{COIN}/\textbf{HASAC}/\textbf{DiCO}) can dominate learning.
This can destabilize teammates and collapse coordination, reducing team return.
\subsection{Ablation Study}\label{subsec:4.2}
\begin{figure}[b!]
    \vskip -0.2in
    \centering
    \subfigure[\textbf{State Factorization.} \newline Effect of agent-centric successor distance.]{
    \includegraphics[width=0.202\textwidth]{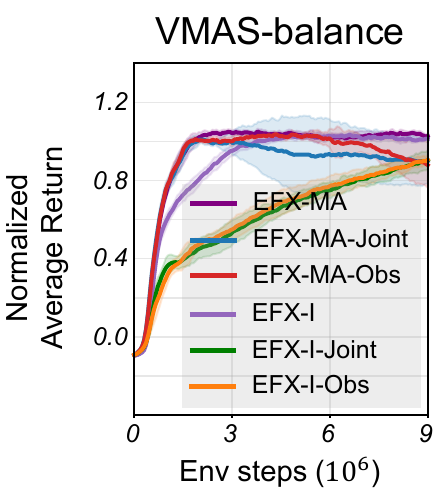}
    \label{fig:ablation_sub1}
    }
    \hspace{0.005\textwidth}
    \subfigure[\textbf{Intrinsic Reward Design.} We ablate SD-based intrinsic~reward.]{
    \includegraphics[width=0.165\textwidth]{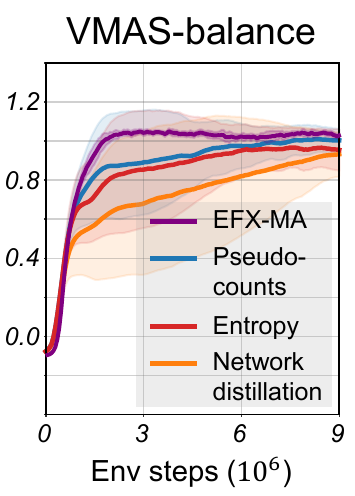}
    \label{fig:ablation_sub2}
    }
    \hspace{0.006\textwidth}
    \subfigure[\textbf{Feasibility Set.} \newline Effect of running reference in Eq.~\eqref{eq:outer_feasible_set}.]{
    \includegraphics[width=0.162\textwidth]{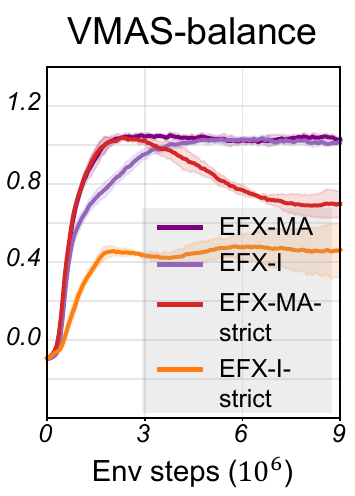}
    \label{fig:ablation_sub3}
    }
    \hspace{0.008\textwidth}
    \subfigure[\textbf{Exploration Budget} $z$. Ablation of fixed budget variants and update rates $\alpha_z$, evaluated only with EFX-MA.]{
    \includegraphics[width=0.324\textwidth]{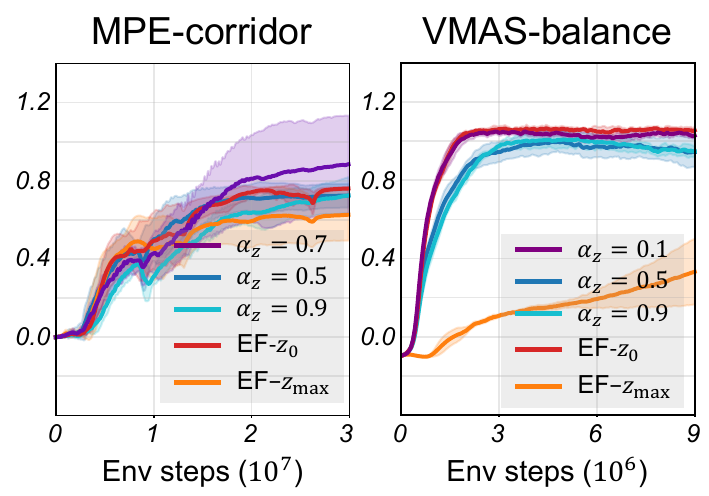}
    \label{fig:ablation_sub4}
    }
    \hfill%
    
    \caption{Ablation study. EFX-MA and EFX-I denote EFXPLORER-MAPPO and EFXPLORER-IPPO. Solid lines show the mean over 5 seeds; shaded regions indicate the standard deviation.}
    \label{fig:ablation}
\end{figure}

We investigate the design choices of \textsc{EFXPLORER} in Fig.~\ref{fig:ablation}, analyzed next:

\noindent\textbf{State Factorization.}
As shown in Fig.~\ref{fig:ablation_sub1}, our default agent-centric novelty (EFX-MA/I) outperforms both joint-state (EFX-MA/I-Joint) and observation-level factorizations (EFX-MA/I-Obs).
Joint-state novelty suffers from high-dimensional estimation, while observation injects teammate non-stationarity that destabilizes SD.
Thus, individual states best balance scalability and stability.

\noindent\textbf{Intrinsic Reward Design.}
Replacing our SD-based reward with alternative bonuses (pseudo-counts~\cite{tang2017exploration}, entropy~\cite{haarnoja2018soft}, network distillation~\cite{burda2018exploration}) degrades performance (Fig.~\ref{fig:ablation_sub2}). 
This highlights the efficacy of long-horizon temporal novelty for discovering coordinated behaviors over short-term exploration signals.

\noindent\textbf{Feasibility Set.}
The default running reference for the feasible set (Eq.~\eqref{eq:outer_feasible_set}) provides smoother budget adaptation.
Conversely, a strict best-return criterion overly restricts budget growth and slows exploration (Fig.~\ref{fig:ablation_sub3}), proving that a moving target avoids being dominated by noisy estimates.

\noindent\textbf{Exploration Budget $z$.}
Fixed-budget variants fail at both extremes (Fig.~\ref{fig:ablation_sub4}): \textsc{EF}-$z_0$ completely suppresses exploration, while \textsc{EF}-$z_{\max}$ over-drives exploration and destabilizes coordination. 
Our adaptive budget resolves this by dynamically matching exploration pressure to task feasibility.
Furthermore, the update rate $\alpha_z$ controls reactivity. 
Smaller $\alpha_z$ (e.g., $0.1$) is more efficient for simpler tasks requiring minimal exploration (\textit{VMAS-balance}), whereas larger $\alpha_z$ (e.g., $0.7$) accelerates discovery in tight bottleneck scenarios (\textit{MPE-corridor}).
\subsection{Efficacy of the Epigraph Formulation}\label{subsec:4.3}
\begin{figure}[t!]
    %\vskip -0.1in
    \begin{center}
        \centerline{\includegraphics[width=1.0\textwidth]{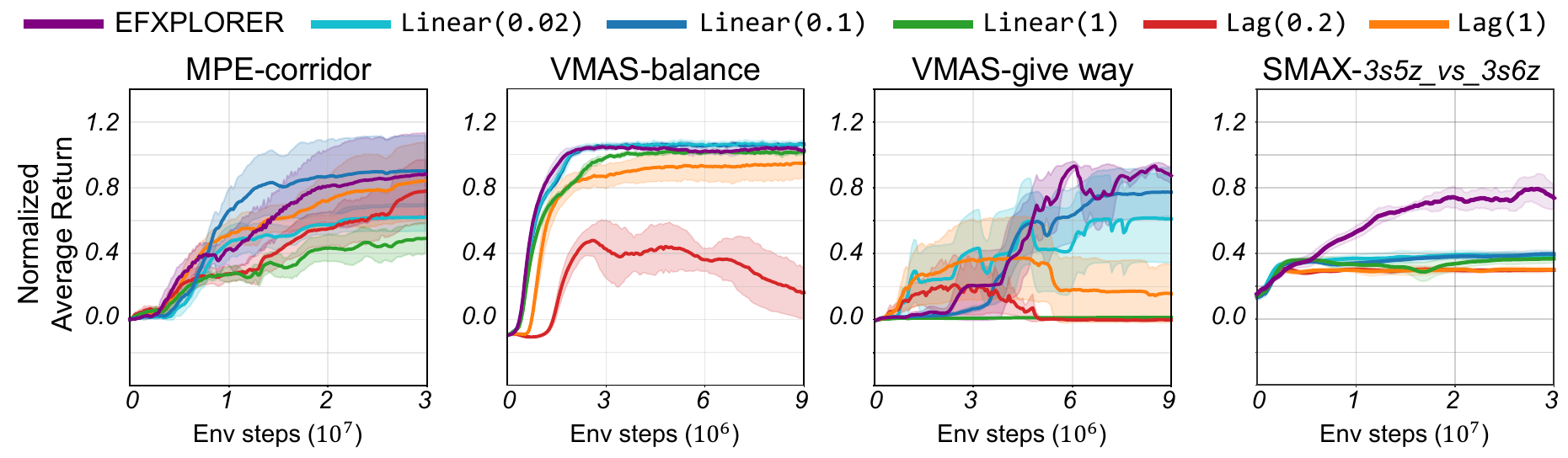}}
        \vskip -0.3in
    \end{center}
    \caption{Performance comparison of \textsc{EFXPLORER} with alternative optimization strategies. Solid lines show the mean across 5 random seeds, while shaded regions indicate the standard deviation. For \textbf{MAPPO-Linear} and \textbf{MAPPO-Lagrangian}, we use \texttt{Linear($\beta$)} and \texttt{Lag($\lambda_0$)} to denote the weight and initialization, respectively.}
    \label{fig:EFXPLORER_results_2}
    \vspace{-10pt}
\end{figure}

To validate our epigraph formulation, we compare \textsc{EFXPLORER} against alternative optimization strategies for solving the constrained objective (Eq.~\eqref{eq:constrained_optimization}).
We consider \textbf{MAPPO-Lagrangian}~\cite{gu2023safe}, which relaxes the hard constraint into a primal-dual objective $\min_{\lambda\geq0}\max_{\boldsymbol{\pi}}\sum J_{i,\text{int}}+\lambda(\Delta J_\text{ext})$. We test initialization values of $\lambda_0=\{0.2,1\}$.
We also evaluate \textbf{MAPPO-Linear}, which optimizes the scalarized objective $J_\text{ext}+\beta \sum J_{i,\text{int}}$ with a weight $\beta=\{0.02,0.1,1\}$.

While \textsc{EFXPLORER} remains consistently competitive across all tasks (Fig.~\ref{fig:EFXPLORER_results_2}), both baselines exhibit severe hyperparameter sensitivity.
\textbf{MAPPO-Linear} is highly brittle to $\beta$; on \textit{MPE-corridor}, $\beta{=}0.1$ yields effective exploration, whereas $0.02$ or $1$ degrade performance. 
Similarly, \textbf{MAPPO-Lagrangian} relies on a heavily weighted initialization ($\lambda_0=1$) to prevent coordination collapse. 
These results highlight the tuning burden of scalarization and Lagrangian relaxations, which \textsc{EFXPLORER} avoids entirely via the epigraph $\min$-bottleneck (further analyzed in Appendix~\ref{sec:appendix_D}).

\begin{figure*}[b!]
    \vskip -0.1in
    \begin{center}
        \centerline{\includegraphics[width=1.0\linewidth]{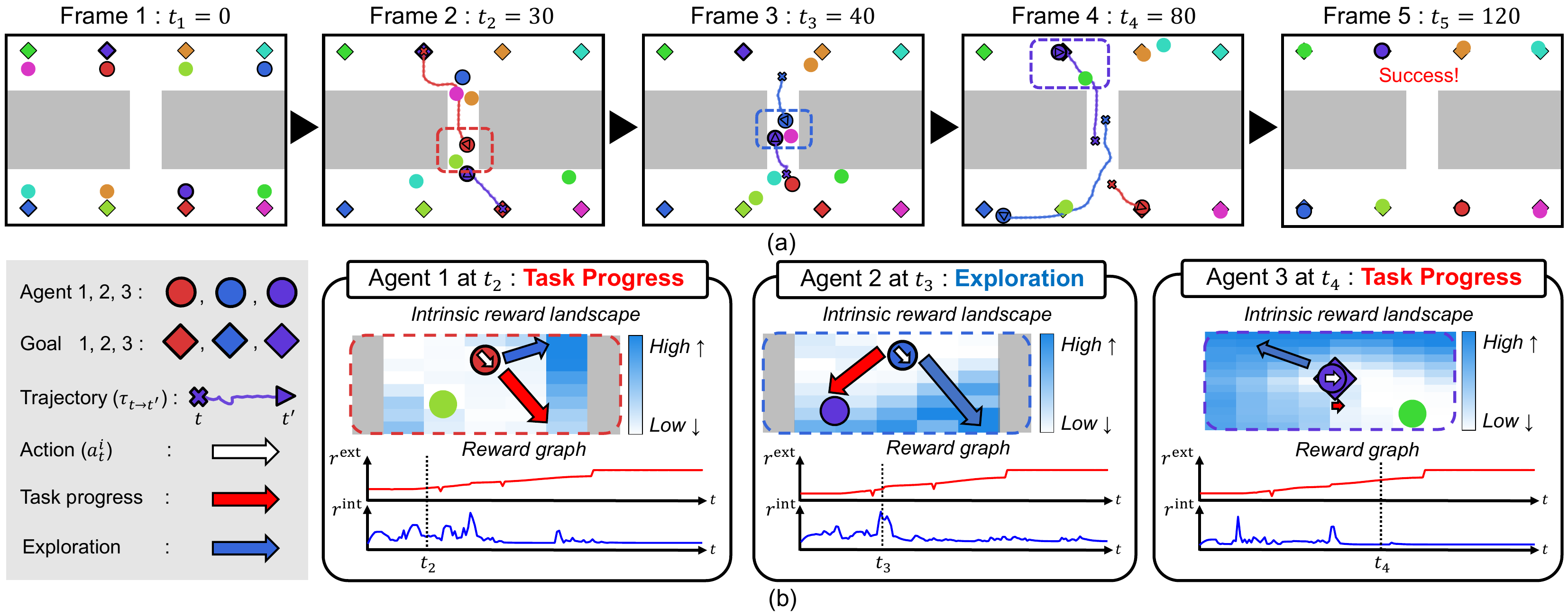}}
        \vskip -0.3in
    \end{center}
    \caption{\textbf{Qualitative analysis on \textit{MPE-corridor}}.
    (a) Episode snapshots ($t_1$--$t_5$) illustrating multi-agent interactions in the narrow passage.
    (b) Stepwise diagnostics for agents 1--3 ($t_2$--$t_4$): intrinsic reward landscapes over candidate next states, action directions maximizing extrinsic or intrinsic reward, executed actions, and corresponding instantaneous reward traces.}
    \label{fig:EFXPLORER_qual_analysis}
    \vspace{-10pt}
\end{figure*}	
\subsection{Case Study: \textit{MPE-corridor}}\label{subsec:4.4}
We use \textit{MPE-corridor} to visualize how \textsc{EFXPLORER} balances exploration and task progress under interaction-induced non-stationarity.
Eight agents (four per side) must pass through a narrow corridor to reach opposite goals.
The extrinsic reward is the $\ell_2$ distance-to-goal, which becomes locally misleading near the corridor entrance.
Thus, purely goal pursuit increases collisions and stalls the team, so exploration is necessary to discover interaction-aware maneuvers.

Fig.~\ref{fig:EFXPLORER_qual_analysis}(a) shows episode snapshots ($t_1$--$t_5$).
Fig.~\ref{fig:EFXPLORER_qual_analysis}(b) analyzes agents~1--3 at representative steps ($t_2$--$t_4$) by visualizing an intrinsic reward landscape over candidate next states, the instantaneous reward traces, and the executed action.
When intrinsic novelty points toward unproductive motion (e.g., toward a wall or away from the goal), \textsc{EFXPLORER} suppresses it and executes a task-aligned action (e.g., $t_2$, $t_4$ at Fig.~\ref{fig:EFXPLORER_qual_analysis}). 
During the corridor bottleneck ($t_3$), it permits a conservative detour to avoid collision.
This detour reduces extrinsic reward while intrinsic reward spikes, suggesting exploration is triggered to resolve the interaction bottleneck.
We further examine whether exploration remains active during temporary reward sacrifice and quantify coordination in SMAX combat scenarios using a local engagement metric in Appendix~\ref{app:exploration_coordination}.
% ================================================================
\section{Related Works}
\label{sec:rel_work}
\subsection{Exploration via Intrinsic Motivation}\label{subsec:5.1}
In cooperative MARL, intrinsic rewards have been widely utilized to alleviate sparse extrinsic rewards.
Early work adapts single-agent curiosity-based methods, such as pseudo-counts~\cite{bellemare2016unifying, tang2017exploration}, prediction errors~\cite{pathak2017curiosity}, and network distillation~\cite{burda2018exploration} to multi-agent settings~\cite{liu2021cooperative, zheng2021episodic}.
Beyond individual curiosity, a significant line of research leverages Mutual Information (MI) to quantify social influence and information gain as intrinsic motivators~\cite{jaques2019social, wang2019influence, jiang2021emergence, li2021celebrating, li2023two}.
However, these methods often rely on additive rewards that are brittle to tune or focus on short-term novelty derived from local prediction errors. 
While recent work~\cite{nimonkar2025self} has explored long-horizon behaviors inspired by goal-conditioned RL~\cite{eysenbach2022contrastive}, explicitly leveraging temporal structures to guide long-term exploration remains underexplored in multi-agent domains.
We address this gap by defining a successor distance-based, factorized episodic novelty as the intrinsic reward to drive long-horizon exploration.
\subsection{Constrained Reinforcement Learning}\label{subsec:5.2}
Constrained RL has primarily been studied in safety-critical settings~\cite{yu2022reachability, gu2024review}, where agents optimize return while satisfying explicit cost or safety constraints, formalized as constrained MDPs~\cite{achiam2017constrained, altman2021constrained}.
A standard solution uses primal-dual updates with Lagrange multipliers to trade off reward and constraint violations, but the induced coupling can be sensitive and may blur the distinction between task objective and constraint \cite{gu2021multi}.
Recent works therefore consider epigraph-based reformulations to separate task optimization from constraint handling~\cite{so2023solving, zhang2025solving}.

Beyond safety constraints, primal-dual machinery can also regulate exploration, e.g., via adaptive entropy regularization~\cite{haarnoja2018soft, kim2023adaptive}.
HASAC~\cite{liu2023maximum} further leverages a maximum-entropy objective to encourage exploration in heterogeneous teams.
Closer to intrinsic motivation, EIPO~\cite{chen2022redeeming} casts intrinsic learning as constrained optimization in the single-agent setting, optimizing a joint objective over extrinsic and intrinsic returns while enforcing an extrinsic-optimality constraint.
Rather than constraining the combined objective, we maximize intrinsic return under a conservative non-degradation constraint on the joint extrinsic return, and extend this view to cooperative MARL. 
We further adopt an epigraph formulation to disentangle exploration incentives from task optimization.
% ================================================================
\section{Conclusions and Future Works}\label{sec:conclusion}
In this paper, we propose \textsc{EFXPLORER}, a conservative exploration framework for cooperative MARL that aims to preserve established coordination.
Its epigraph-based $\min$-coupling keeps task and exploration signals separate and regulates exploration according to task feasibility.
Our experiments show that \textsc{EFXPLORER} supports the progressive discovery of new cooperative behavior patterns while maintaining task performance.

\paragraph{Future Work.} 
Our current implementation relies on policy gradients and synchronized updates from a common reference policy.
Extending it to value-based MARL~\cite{sunehag2017value, rashid2020monotonic} would require a suitable decomposition of the joint extrinsic advantage.
To relax the reliance on a common reference policy, future work could support sequential update schemes such as HASAC~\cite{liu2023maximum}.
This could involve intermediate joint reference~\cite{kuba2021trust} and off-policy corrections~\cite{wang2023order}.
In addition, our exploration mechanism uses agent-centric successor distances to track novelty for each agent.
Graph-conditioned representations could complement this approach by capturing cross-agent novelty and supporting exploration in larger teams.

Our evaluation is limited to state-based observations, and scaling \textsc{EFXPLORER} to high-dimensional observations (e.g., raw images) remains an important next step. 
Furthermore, a promising direction is broadening the epigraph formulation to jointly manage exploration, safety constraints, and human preferences within a unified framework.
For example, RL-based LLM post-training could use this formulation to promote response diversity under performance and alignment constraints.

\section*{Acknowledgments}
We would like to thank the anonymous reviewers for their insightful comments.
This work was partly supported by Samsung Research Funding \& Incubation Center of Samsung Electronics under Project Number SRFC-IT2402-17, Institute of Information \& communications Technology Planning \& Evaluation (IITP) grant funded by the Korean government (MSIT) [NO.RS-2021-II211343, Artificial Intelligence Graduate School Program (Seoul National University)] and Hyundai Motor Chung Mong-Koo Foundation.

\newpage
\bibliographystyle{plainnat}
\bibliography{neurips_2026}

%%%%%%%%%%%%%%%%%%%%%%%%%%%%%%%%%%%%%%%%%%%%%%%%%%%%%%%%%%%%%%%%%%%%%%%%%%%%%%%
%%%%%%%%%%%%%%%%%%%%%%%%%%%%%%%%%%%%%%%%%%%%%%%%%%%%%%%%%%%%%%%%%%%%%%%%%%%%%%%
% APPENDIX
%%%%%%%%%%%%%%%%%%%%%%%%%%%%%%%%%%%%%%%%%%%%%%%%%%%%%%%%%%%%%%%%%%%%%%%%%%%%%%%
%%%%%%%%%%%%%%%%%%%%%%%%%%%%%%%%%%%%%%%%%%%%%%%%%%%%%%%%%%%%%%%%%%%%%%%%%%%%%%%
\newpage
\appendix
\onecolumn

% ToC of Appendix
\let\addcontentsline\oldaddcontentsline
\renewcommand{\contentsname}{\Large{Appendix}}
\setcounter{tocdepth}{4}
\tableofcontents
\clearpage

\section{Theoretical Details and Proof}\label{sec:appendix_A}
In this section, we provide the theoretical details deferred from the main paper.
In particular, we (i) establish the foundations of Sec.~\ref{subsec:3.1}, prove Proposition~\ref{prop:epi_surr} and Theorem~\ref{thm:epi_pg}, and discuss the approximation error of the fixed-reference return surrogate, and (ii) provide theoretical results successor-distance learning (Sec.~\ref{subsec:3.2}).

\subsection{Conservative Exploration via Epigraph Reformulation}\label{subsec:Appendix_epi}
This section provides theoretical foundations for the conservative exploration framework introduced in Sec.~\ref{subsec:3.1}.
We first formalize the synchronized joint-update assumption, connecting our constrained objective to standard MARL training paradigms.
We then derive a conservative stepwise sufficient condition for the trajectory-level task constraint $J_{\text{ext}}(\boldsymbol{\pi})\geq J_{\text{ext}}(\hat{\boldsymbol{\pi}})$ using the stepwise constraints based on the reference advantage $A^\text{ext}_{\hat{\boldsymbol{\pi}}}(s,\mathbf{a})$ via the performance difference identity~\cite{kakade2002approximately, schulman2015trust}.
Next, we define the budget dynamics and the associated epigraph Bellman backup operator and prove Proposition~\ref{prop:epi_surr}.
We then derive Theorem~\ref{thm:epi_pg}, which justifies using the epigraph advantage in actor--critic updates.
Finally, we analyze the approximation error of the fixed-reference return surrogate.

\subsubsection{Justification for Synchronized Joint Updates}\label{app:epi_synchronized}
We rely on a \textbf{synchronized joint-update assumption}: all agents share a single candidate joint policy $\boldsymbol{\pi}$ around a shared reference $\hat{\boldsymbol{\pi}}$, rather than optimizing over separate hypothetical joint policies. 

Decoupling constraints per agent would misalign gradient directions and collapse coordination. 
Fortunately, this assumption naturally aligns with standard CTDE frameworks (e.g., MAPPO~\cite{yu2022surprising}), where agents collect data synchronously and update policies simultaneously. 
Bounded by trust-region mechanisms (e.g., PPO clipping), agents safely and collectively step toward a unified new joint policy.

A limitation of this synchronized formulation is that it does not directly accommodate asynchronous or sequential update schemes, such as Heterogeneous-Agent Soft Actor-Critic (HASAC), where agents update their policies one by one~\cite{liu2023maximum}.
In such settings, the joint reference policy shifts continuously during a single training iteration, which fundamentally destabilizes the shared reference. 
Extending the epigraph bottleneck to handle sequential agent-wise updates without a synchronized joint reference is an important direction for future work.
For example, sequential updates could use stage-wise references with policy-ratio corrections~\cite{kuba2021trust} or truncated importance weights~\cite{wang2023order} to address the resulting off-policy mismatch.

\subsubsection{Stepwise Feasibility from the Performance Difference Identity}\label{app:epi_stepwise}
We justify the use of the reference extrinsic advantage $A^\text{ext}_{\hat{\boldsymbol{\pi}}}(s, \mathbf{a})$ as a stepwise surrogate signal for the trajectory-level task constraint $J_\text{ext}(\boldsymbol{\pi})\geq J_\text{ext}(\hat{\boldsymbol{\pi}})$.
The key tool is the advantage-based performance difference identity~\cite{schulman2015trust}.

Consider the Dec-POMDP~\cite{oliehoek2016concise} induced by the joint policy $\boldsymbol{\pi}$ with extrinsic return $J_\text{ext}(\boldsymbol{\pi})\triangleq\mathbb{E}_{\tau\sim\boldsymbol{\pi}}\big[\sum_{t\geq0}\gamma^t r_t^\text{ext}\big]$.
Let $V^{\text{ext}}_{\hat{\boldsymbol{\pi}}}$ and $Q^{\text{ext}}_{\hat{\boldsymbol{\pi}}}$ be the value functions of the reference joint policy
$\hat{\boldsymbol{\pi}}$, and define the reference advantage
$A^{\text{ext}}_{\hat{\boldsymbol{\pi}}}(s,\mathbf{a})
=Q^{\text{ext}}_{\hat{\boldsymbol{\pi}}}(s,\mathbf{a})-V^{\text{ext}}_{\hat{\boldsymbol{\pi}}}(s)$.
Let $d^{\boldsymbol{\pi}}(s) \triangleq (1-\gamma)\sum_{t\ge0}\gamma^t \Pr_{\boldsymbol{\pi}}(s_t=s)$
denote the discounted state-visitation distribution under $\boldsymbol{\pi}$.
Then the following identity holds:
\begin{equation}
    J_{\text{ext}}(\boldsymbol{\pi})-J_{\text{ext}}(\hat{\boldsymbol{\pi}})
    =\frac{1}{1-\gamma}\;\mathbb{E}_{s\sim d^{\boldsymbol{\pi}},\ \mathbf{a}\sim \boldsymbol{\pi}(\cdot|s)}
    \big[A^{\text{ext}}_{\hat{\boldsymbol{\pi}}}(s,\mathbf{a})\big].
\label{eq:perf_diff_lemma}
\end{equation}
While the original constraint only requires the sum of expected advantages (Eq.~\eqref{eq:perf_diff_lemma}) to be non-negative, it allows for temporary performance degradation along a trajectory.
We therefore consider a stricter, pointwise sufficient condition for non-degradation to encourage monotone task improvement during exploration.

\begin{theorem}[Sufficiency of stepwise non-negativity]\label{thm:suff_stepwise}
If $A^{\text{ext}}_{\hat{\boldsymbol{\pi}}}(s,\mathbf{a}) \ge 0$ for all states $s$ and joint actions $\mathbf{a}\in\text{supp}(\boldsymbol{\pi}(\cdot|s))$,
then $J_{\text{ext}}(\boldsymbol{\pi}) \ge J_{\text{ext}}(\hat{\boldsymbol{\pi}})$.
\end{theorem}

\noindent\textit{Proof.}
By Eq.~\eqref{eq:perf_diff_lemma},
$J_{\text{ext}}(\boldsymbol{\pi})-J_{\text{ext}}(\hat{\boldsymbol{\pi}})$
is proportional to an expectation of $A^{\text{ext}}_{\hat{\boldsymbol{\pi}}}(s,\mathbf{a})$
under nonnegative weights $d^{\boldsymbol{\pi}}(s)\boldsymbol{\pi}(\mathbf{a}|s)$.
If $A^\text{ext}_{\hat{\boldsymbol{\pi}}}(s,\mathbf{a})\geq 0$ for all $(s,\mathbf{a})$ in the support of $\boldsymbol{\pi}$, the expectation is nonnegative.
\hfill$\square$

We adopt this conservative stepwise view to mitigate exploration-induced distribution shift.
Practically, this stepwise surrogate also aligns with implementation: our Epigraph-GAE estimates a \textit{per-timestep} bottleneck signal, thereby enabling feasibility checks without explicitly estimating trajectory-level improvement.
\subsubsection{Proof of Proposition~\ref{prop:epi_surr}}\label{app:prop_epi_surr}

\noindent\textbf{Proposition~\ref{prop:epi_surr}} (Conservative Epigraph Feasibility).

\textit{Enforcing the condition $\mathcal{B}_i^\mathrm{epi}(\tilde{s}_t^i,\mathbf{a}_t)\ge0$ for all visited states and actions provides a conservative sufficient condition for feasibility of the original epigraph constraints.}

To prove Proposition~\ref{prop:epi_surr}, we first establish a lemma regarding the recursive structure of the intrinsic surplus $S_i^\text{int}$.

\begin{lemma}[Recursive structure of intrinsic surplus]\label{lemma:recurr_surplus}
Given the budget dynamics $\gamma z_{t+1}^i=z_t^i-r_{i,t}^\mathrm{int}$, the intrinsic surplus $ S_i^\mathrm{int}(\tilde{s}_t^i, \mathbf{a}_t)$ satisfies the following recursive relation:
\begin{equation}
    S_i^\mathrm{int}(\tilde{s}_t^i, \mathbf{a}_t) = \gamma \mathbb{E}\Big[ S_i^\mathrm{int}(\tilde{s}_{t+1}^i, \mathbf{a}_{t+1}) \mid \tilde{s}_t^i, \mathbf{a}_t \Big].
    \label{eq:lemma_surplus_recursion}
\end{equation}
This implies that $S_i^\mathrm{int}$ behaves as a valid value function representing the expected future budget surplus.
\end{lemma}

\begin{proof}
Recall the definition of the intrinsic surplus in Eq.~\eqref{eq:surplus_def}:
\begin{equation}
    S_i^\text{int}(\tilde{s}_t^i, \mathbf{a}_t) \triangleq \mathbb{E}_{\tau}\left[ \sum_{k=t}^{\infty} \gamma^{k-t} r_{i,k}^\text{int} - z_t^i \middle|
    \tilde{s}_t^i, \mathbf{a}_t \right].
\end{equation}
Now, consider the expectation of the surplus at the next timestep $t+1$:
\begin{equation}
    \mathbb{E}[S_i^\text{int}(\tilde{s}_{t+1}^i, \mathbf{a}_{t+1}) \mid \tilde{s}_t^i, \mathbf{a}_t] = \mathbb{E}\left[ \sum_{k=t+1}^{\infty} \gamma^{k-(t+1)} r_{i,k}^\text{int} - z_{t+1}^i | \tilde{s}_t^i, \mathbf{a}_t \right].
\end{equation}
Multiplying by $\gamma$ and substituting the budget dynamics ($\gamma z_{t+1}^i = z_t^i - r_{i,t}^\text{int}$):
\begin{align}
    \gamma \mathbb{E}[S_i^\text{int}(\tilde{s}_{t+1}^i, \mathbf{a}_{t+1})]
    &= \mathbb{E}\left[ \sum_{k=t+1}^{\infty} \gamma^{k-t} r_{i,k}^\text{int} - \gamma z_{t+1}^i \right] \notag \\
    &= \mathbb{E}\left[ \sum_{k=t+1}^{\infty} \gamma^{k-t} r_{i,k}^\text{int} - (z_t^i - r_{i,t}^\text{int}) \right] \notag \\
    &= \mathbb{E}\left[ \sum_{k=t}^{\infty} \gamma^{k-t} r_{i,k}^\text{int} - z_t^i \right] \notag \\
    &= S_i^\text{int}(\tilde{s}_t^i, \mathbf{a}_t).
\end{align}
This confirms the recursive property.
\end{proof}

\textbf{Proof of Proposition~\ref{prop:epi_surr}.} 
Assume the stepwise epigraph bottleneck is nonnegative along trajectories induced by $\boldsymbol{\pi}$: 
\begin{equation}
    \forall i\in\mathcal{N},\ \forall \tau\sim\boldsymbol{\pi}, \ \forall t\geq0:\quad 
    \mathcal{B}_i^{\text{epi}}(\tilde{s}_t^i,\mathbf{a}_t) =
    \min\Big\{
    A_{\hat{\boldsymbol{\pi}}}^\text{ext}(s_t, \mathbf{a}_t), S_i^\text{int}(\tilde{s}_t^i,\mathbf{a}_t)
    \Big\}\geq 0.
    \label{eq:epi_stepwise_assumption}
\end{equation} 

By definition of the minimum operator, Eq.~\eqref{eq:epi_stepwise_assumption} implies that along the same trajectories: 
\begin{equation}
    \forall i\in\mathcal{N},\ \forall \tau\sim\boldsymbol{\pi}, \ \forall t \geq 0: \quad A_{\hat{\boldsymbol{\pi}}}^{\text{ext}}(s_t,\mathbf{a}_t) \ge 0 \quad \text{and} \quad S_{i}^{\text{int}}(\tilde{s}_t^i,\mathbf{a}_t) \ge 0.
    \label{eq:both_nonneg}
\end{equation}

\textbf{Extrinsic Feasibility:} Since $A_{\hat{\boldsymbol{\pi}}}^\text{ext}(s_t,\mathbf{a}_t)\ge 0$ for all $t$ along $\tau\sim\boldsymbol{\pi}$, we have:
\begin{equation}
    \mathbb{E}_{\tau\sim\boldsymbol{\pi}}
    \left[\sum_{t=0}^{\infty}\gamma^t A_{\hat{\boldsymbol{\pi}}}^\text{ext}(s_t,\mathbf{a}_t)\right]\ge 0.
\end{equation}
By the performance difference identity (Eq.~\eqref{eq:perf_diff_lemma}),
this yields $\Delta J_\text{ext}(\boldsymbol{\pi})\ge 0$, hence
\begin{equation}
    J_\text{ext}(\boldsymbol{\pi})\ge J_\text{ext}(\hat{\boldsymbol{\pi}}).
\end{equation}

\textbf{Intrinsic Feasibility:} From Eq.~\eqref{eq:both_nonneg},  $S_i^\text{int}(\tilde{s}_0^i,\mathbf{a}_0)\ge 0$ holds for all $\tau\sim\boldsymbol{\pi}$.
Taking expectation over $\tau\sim\boldsymbol{\pi}$ gives:
\begin{align}
\mathbb{E}_{\tau\sim\boldsymbol{\pi}}\!\left[S_i^\text{int}(\tilde{s}_0^i,\mathbf{a}_0)\right]
&=
\mathbb{E}_{\tau\sim\boldsymbol{\pi}}
\left[\sum_{k=0}^{\infty}\gamma^k r_{i,k}^\text{int} - z_0^i\right] \\
&=
\mathbb{E}_{\tau\sim\boldsymbol{\pi}}
\left[\sum_{k=0}^{\infty}\gamma^k r_{i,k}^\text{int}\right]-z^i\\
&=
J_{i,\text{int}}(\pi^i)-z^i\geq0,
\end{align}
where we used $z_0^i=z^i$.
Therefore $J_{i,\text{int}}(\pi^i)\ge z^i$.
Lemma~\ref{lemma:recurr_surplus} further shows that $S_i^\text{int}$ admits a valid continuation value under the budget dynamics.

Combining the two parts, the step-wise condition $\mathcal{B}_i^{\text{epi}}(\tilde{s}_t^i,\mathbf{a}_t)\ge 0$
is a conservative sufficient condition for satisfying both
$J_\text{ext}(\boldsymbol{\pi})\ge J_\text{ext}(\hat{\boldsymbol{\pi}})$ and $J_{i,\text{int}}(\pi^i)\ge z^i$,
and hence for feasibility of the epigraph constraints in Eq.~\eqref{eq:epigraph_form}.
\qed
\subsubsection{Proof of Theorem~\ref{thm:epi_pg}}\label{app:epi_pg}

While Proposition~\ref{prop:epi_surr} provides a pointwise guarantee, policy optimization requires a scalar objective.
By Jensen's inequality on the concave minimum operator, the expected bottleneck lower-bounds the true feasibility margins:
\begin{equation}
    \mathbb{E}\big[\min(A_{\hat{\boldsymbol{\pi}}}^\text{ext}, S_i^\text{int})\big] \le \min\big(\mathbb{E}[A_{\hat{\boldsymbol{\pi}}}^\text{ext}], \mathbb{E}[S_i^\text{int}]\big).
\end{equation}
Thus, the expected cumulative bottleneck serves as a provably conservative global objective for trajectory-level feasibility:
\begin{equation}
    J_i^\mathrm{epi}(\boldsymbol{\pi}) \triangleq \mathbb{E}_{\tau} \left[ \sum_{t=0}^{\infty} \gamma^t \mathcal{B}_{i}^{\mathrm{epi}}(\tilde{s}_t^i, \mathbf{a}_t) \right].
\end{equation}

Directly differentiating $J_i^\mathrm{epi}(\boldsymbol{\pi})$ is intractable due to policy-induced distribution shifts.
We circumvent this using a local fixed-reference surrogate~\cite{schulman2015trust}, which treats the bottleneck as a constant evaluated under the reference policy $\hat{\boldsymbol{\pi}}$.

\textbf{Theorem~\ref{thm:epi_pg}} (Local Epigraph Policy Gradient).
\textit{Let $\pi_\theta^i$ be the parameterized policy of agent $i$, and let $\hat{\boldsymbol{\pi}}$ be the fixed reference policy. 
The gradient of the local epigraph surrogate $L_i^\mathrm{epi}(\theta)$ is:
\begin{equation}
    \nabla_\theta L_i^{\mathrm{epi}}(\theta) = \hat{\mathbb{E}}_t \left[ \rho^i_t(\theta) \nabla_\theta \log \pi_\theta^i(a_t^i|o_t^i) A^{\mathrm{epi},\hat{\boldsymbol{\pi}}}_{i,t} \right],
    \label{eq:epi_pg}
\end{equation}
where $A_{i,t}^{\mathrm{epi},\hat{\boldsymbol{\pi}}}\triangleq \mathcal{B}_{i}^{\mathrm{epi}}(\tilde{s}_t^i, \mathbf{a}_t)-b_{i}(\tilde{s}_t^i)$ for any action-independent baseline $b_{i}(\tilde{s}_t^i)$.}

\begin{proof}
We define the local epigraph surrogate objective using the fixed reference policy $\hat{\boldsymbol{\pi}}$ to evaluate the bottleneck:
\begin{equation}
    L_i^{\mathrm{epi}}(\theta) = \hat{\mathbb{E}}_t \left[ \frac{\pi_\theta^i(a_t^i|o_t^i)}{\hat{\pi}^i(a_t^i|o_t^i)} \mathcal{B}_{i}^{\mathrm{epi}}(\tilde{s}_t^i,\mathbf{a}_t) \right].
\end{equation}

Since the surrogate's distribution and value terms ($A_{\hat{\boldsymbol{\pi}}}^{\text{ext}}$, $S_i^{\text{int}}$) are fixed to the reference policy~\cite{schulman2015trust}, $\mathcal{B}_{i}^{\mathrm{epi}}$ is a constant with respect to $\theta$.
Taking the gradient of $L_i^{\mathrm{epi}}(\theta)$ with respect to $\theta$ yields:
\begin{equation}
    \nabla_\theta L_i^{\mathrm{epi}}(\theta) = \hat{\mathbb{E}}_t \left[ \frac{\nabla_\theta \pi_\theta^i(a_t^i|o_t^i)}{\hat{\pi}^i(a_t^i|o_t^i)} \mathcal{B}_{i}^{\mathrm{epi}}(\tilde{s}_t^i,\mathbf{a}_t) \right].
\end{equation}

Applying the log-derivative trick, $\nabla_\theta \pi_\theta^i = \pi_\theta^i \nabla_\theta \log \pi_\theta^i$, we obtain:
\begin{align}
    \nabla_\theta L_i^{\mathrm{epi}}(\theta) 
    &= \hat{\mathbb{E}}_t \left[ \frac{\pi_\theta^i(a_t^i|o_t^i)}{\hat{\pi}^i(a_t^i|o_t^i)} \nabla_\theta \log \pi_\theta^i(a_t^i|o_t^i) \mathcal{B}_{i}^{\mathrm{epi}}(\tilde{s}_t^i,\mathbf{a}_t) \right] \notag \\
    &= \hat{\mathbb{E}}_t \left[ \rho^i_t(\theta) \nabla_\theta \log \pi_\theta^i(a_t^i|o_t^i) \mathcal{B}_{i}^{\mathrm{epi}}(\tilde{s}_t^i,\mathbf{a}_t) \right].
\end{align}

As in standard policy gradient methods, subtracting an action-independent baseline $b_i(\tilde{s}_t^i)$ leaves the expected gradient unchanged. 
Defining the epigraph advantage as $A^{\mathrm{epi},\hat{\boldsymbol{\pi}}}_{i,t} = \mathcal{B}_{i}^{\mathrm{epi}}(\tilde{s}_t^i,\mathbf{a}_t) - b_i(\tilde{s}_t^i)$, we recover the final gradient expression.
\end{proof}
\subsubsection{Approximation Error of the Local Surrogate}\label{app:app_err_local_surrogate}

Section~\ref{subsec:3.1} motivates local surrogate optimization using the performance difference identity in Eq.~\eqref{eq:performance_difference}.
This identity involves the candidate joint policy's state-visitation distribution, whereas the local surrogate uses trajectories collected under the reference policy.
We quantify the resulting approximation error in extrinsic return when multiple agents update their policies jointly.

For a reference joint policy $\hat{\boldsymbol{\pi}}$ and a candidate joint policy $\boldsymbol{\pi}$, define the fixed-reference return surrogate as:
\begin{equation}
    \tilde{J}_{\hat{\boldsymbol{\pi}}}^{\text{ext}}(\boldsymbol{\pi})=J_\text{ext}(\hat{\boldsymbol{\pi}})+\frac{1}{1-\gamma}\mathbb{E}_{s\sim d^{\hat{\boldsymbol{\pi}}}, \mathbf{a}\sim\boldsymbol{\pi}}[A_{\hat{\boldsymbol{\pi}}}^\text{ext}(s, \mathbf{a})],
\end{equation}
where $d^{\hat{\boldsymbol{\pi}}}$ is the normalized discounted state-visitation distribution.
This surrogate retains the candidate action distribution while fixing the state distribution to the reference policy.

Assume factorized joint policies and bounded reference advantages.
Let
\begin{equation}
    \epsilon = \sup_{s, \mathbf{a}} |A_{\hat{\boldsymbol{\pi}}}^{\text{ext}}(s, \mathbf{a})|, \quad \alpha_i=D_{\text{TV}}^{\text{max}} (\hat{\pi}^i, \pi^i),
\end{equation}
where $D_{\text{TV}}^{\text{max}}$ denotes the maximum total variation distance over policy inputs.
Applying the multi-agent coupling argument~\cite{wu2021coordinated} gives:
\begin{equation}
    |J_\text{ext}(\boldsymbol{\pi})-\tilde{J}_{\hat{\boldsymbol{\pi}}}^\text{ext}(\boldsymbol{\pi})|\leq 4\epsilon \big[\frac{1-\gamma\prod_{i=1}^N(1-\alpha_i)}{1-\gamma}-1\big].
\end{equation}

Letting $\alpha=\max_i \alpha_i$, the inequality $1-\prod_i(1-\alpha_i)\leq \sum_i \alpha_i \leq N\alpha$ yields:
\begin{equation}
    |J_\text{ext}(\boldsymbol{\pi})-\tilde{J}_{\hat{\boldsymbol{\pi}}}^\text{ext}(\boldsymbol{\pi})|\leq \frac{4\epsilon\gamma}{1-\gamma}N\alpha.
\end{equation}
For fixed $\gamma$, this gives $O(\epsilon N \alpha)$ upper bound.
It highlights the importance of limiting individual policy changes when using a fixed reference, particularly as the number of agents increases.
This bound quantifies the error from fixing the state-visitation distribution in the extrinsic return surrogate, assuming exact reference advantages.
It does not cover estimation errors or the remaining approximations in the practical epigraph update.
\subsection{Theoretical Analysis of Factorized Successor Distance}\label{subsec:Appendix_SD}
This section provides proofs for the convergence of the learned distance metric and its recovery of the marginal successor distance, building on the proof in~\cite{myers2024learning, myers2025offline}.

\subsubsection{Convergence on Quasimetric Structure over $\mathcal{S}_i$}

We define the space of non-negative functions on $\mathcal{S}_i$ as $\mathcal{D}_i \triangleq \{d:\mathcal{S}_i \times \mathcal{S}_i \rightarrow \mathbb{R}_{\ge 0}\mid d(x,x)=0, \forall x\}$.
We enforce $d_{\phi_i}$ to satisfy the quasimetric axioms via the set $\mathcal{Q}_i$:
\begin{equation}
    \mathcal{Q}_i \triangleq \{d \in \mathcal{D}_i : d(x, z) \le d(x, y) + d(y, z), \quad \forall x, y, z \in \mathcal{S}_i\}.
\end{equation}
We utilize the path relaxation operator $\mathcal{K}: \mathcal{D}_i \to \mathcal{D}_i$ defined with the infimum:
\begin{equation}
    \mathcal{K}(d)(x, z) \triangleq \inf_{y \in \mathcal{S}_i} [d(x, y) + d(y, z)].
\end{equation}
Note that $\mathcal K$ maps $\mathcal D_i$ to itself: for any $d\in\mathcal D_i$, $\mathcal{K}(d)(x,x)=\inf_y[d(x,y)+d(y,x)]\le d(x,x)+d(x,x)=0$, and since $\mathcal{K}(d) \ge 0$, we have $\mathcal{K}(d)(x,x)=0$.

\begin{lemma}
    \label{lem:quasimetric_fixed}
    A distance $d \in \mathcal{D}_i$ is a quasimetric ($d \in \mathcal{Q}_i$) if and only if it is a fixed point of the path relaxation operator ($\mathcal{K}(d) = d$).
\end{lemma}
\begin{proof}
    ($\Rightarrow$) If $d \in \mathcal{Q}_i$, then $d(x, z) \le d(x, y) + d(y, z)$ for all $y$. Taking the infimum over $y$ yields $d(x, z) \le \mathcal{K}(d)(x, z)$. Conversely, choosing $y=x$ implies $\mathcal{K}(d)(x, z) \le d(x, x) + d(x, z) = d(x, z)$. Thus, $\mathcal{K}(d) = d$.
    
    \noindent ($\Leftarrow$) If $\mathcal{K}(d) = d$, then for any $y$, $d(x, z) = \inf_{w}[d(x, w) + d(w, z)] \le d(x, y) + d(y, z)$. This satisfies the triangle inequality.
\end{proof}

\begin{lemma}
    \label{lem:order-preserving}
    If $d \le d'$ pointwise, then $\mathcal K(d) \le \mathcal K(d')$ pointwise.
\end{lemma}
\begin{proof}
    For any $x,z$ and any $y$, $d(x,y)+d(y,z)\le d'(x,y)+d'(y,z)$.
    Taking infimum over $y$ yields $\mathcal K(d)(x,z)\le \mathcal K(d')(x,z)$.
\end{proof}

\begin{theorem}[Convergence]
    \label{thm:path_convergence}
    For any initial distance $d_0 \in \mathcal{D}_i$, the sequence defined by $d_{n+1} = \mathcal{K}(d_n)$ converges pointwise to a quasimetric $d_\infty \in \mathcal{Q}_i$.
\end{theorem}
\begin{proof}
    The sequence is monotonically non-increasing: $d_{n+1}(x, z) \le d_n(x, x) + d_n(x, z) = d_n(x, z)$. Since it is lower-bounded by 0, it converges pointwise to a limit $d_\infty$. We show $d_\infty$ is a fixed point of $\mathcal{K}$.
    
    First, by Lemma~\ref{lem:order-preserving}, $\mathcal{K}$, $d_\infty \le d_n \implies \mathcal{K}(d_\infty) \le \mathcal{K}(d_n) = d_{n+1}$. 
    Taking the limit $n \to \infty$, we get $\mathcal{K}(d_\infty) \le d_\infty$.
    
    Second, we show $d_\infty \le \mathcal{K}(d_\infty)$. 
    Fix $x, z$ and define $f_n(y) \triangleq d_n(x, y) + d_n(y, z)$.
    Note that $f_n(y)$ converges pointwise to $f_\infty(y) \triangleq d_\infty(x, y) + d_\infty(y, z)$.
    For any $\epsilon > 0$, there exists $y_\epsilon$ such that $f_\infty(y_\epsilon) \le \inf_y f_\infty(y) + \epsilon = \mathcal{K}(d_\infty)(x, z) + \epsilon$.
    Since $d_{n+1}(x, z) = \inf_y f_n(y) \le f_n(y_\epsilon)$, taking the limit yields:
    \begin{equation}
        d_\infty(x, z) \le f_\infty(y_\epsilon) \le \mathcal{K}(d_\infty)(x, z) + \epsilon.
    \end{equation}
    Since $\epsilon$ is arbitrary, $d_\infty \le \mathcal{K}(d_\infty)$.
    Combined with $\mathcal{K}(d_\infty) \le d_\infty$, we have $\mathcal{K}(d_\infty) = d_\infty$. 
    By Lemma~\ref{lem:quasimetric_fixed}, $d_\infty \in \mathcal{Q}_i$.
\end{proof}
\subsubsection{Recovery of Marginal Successor Distance}
Following the contrastive metric distillation framework of \cite{myers2024learning}, we show local InfoNCE recovers a temporal proximity consistent with the marginal discounted successor occupancy under the joint policy $\boldsymbol{\pi}$.
As formalized in Section~\ref{subsec:2.1}, the global state decomposes as $s_t=[s_t^1, \dots, s_t^n, s_t^{\text{env}}]$.
For agent $i$, denote the complement components by $s_t^{-i}$.
Under the joint policy $\mathbf{a}_t \sim \boldsymbol{\pi}(\cdot\mid s_t)$ and environment dynamics $\mathcal{P}$, the induced marginal transition on $\mathcal{S}_i$ is:
\begin{equation}
    P^{\boldsymbol{\pi}}(s_{t+1}^i \mid s_t^i)
    =\sum_{s_t^{-i},\,\mathbf{a}_t}
    \mathcal{P}(s_{t+1}^i \mid s_t^i, s_t^{-i}, \mathbf{a}_t)\,
    \boldsymbol{\pi}(\mathbf{a}_t \mid s_t^i, s_t^{-i})\,
    P^{\boldsymbol{\pi}}_t(s_t^{-i} \mid s_t^i),
    \label{eq:marginal_transition}
\end{equation}
where $P^{\boldsymbol{\pi}}_t(s_t^{-i} \mid s_t^i)$ is the time-$t$ conditional distribution induced by $\boldsymbol{\pi}$~\cite{myers2024empowerment}.

Define the marginal discounted successor occupancy measure:
\begin{equation}
    M^{\boldsymbol{\pi}}_i(y\mid x) \triangleq (1-\gamma)\sum_{t'=0}^\infty \gamma^{t'}\Pr(s_{t+t'}^i=y \mid s_t^i=x,\boldsymbol{\pi}),
    \label{eq:marginal_successor}
\end{equation}
where transitions follow Eq.~\eqref{eq:marginal_transition}.
Equivalently~\cite{myers2024learning}, sampling $t'\sim\mathrm{Geom}(1-\gamma)$ and setting $(x,y)=(s_t^i,s_{t+t'}^i)$ yields $y\sim M_i^{\boldsymbol{\pi}}(\cdot\mid x)$.

Let $\rho_i^{\boldsymbol{\pi}}$ be the negative distribution on $\mathcal{S}_i$ (e.g., the replay buffer marginal under $\boldsymbol{\pi}$).
Consider the InfoNCE loss:
\begin{equation}
    \mathcal{L}_{\mathrm{NCE}}(f_i) \triangleq
    \mathbb{E}\Bigg[
    -\log\frac{\exp(f_i(x,y^+))}{\exp(f_i(x,y^+))+\sum_{k=1}^K \exp(f_i(x,y_k^-))}\Bigg],
    \label{eq:local_infonce}
\end{equation}
where $x\sim\mu_i$, $y^+\sim M_i^{\boldsymbol{\pi}}(\cdot\mid x)$, and $y_k^-\sim\rho_i^{\boldsymbol{\pi}}$.
Our InfoNCE loss can be viewed as the symmetric mini-batch analogue of Eq.~\eqref{eq:local_infonce} with $K=B-1$: the first term uses in-batch negatives $\{y_k\}_{k\neq j}$ for each anchor $x_j$, and the second swaps the roles of $x$ and $y$; we idealize these in-batch negatives as i.i.d.\ samples from $\rho_i^{\boldsymbol{\pi}}$.

\begin{proposition}[Optimal critic recovers a density ratio]
\label{prop:nce_density_ratio}
Assume $f_i$ is unconstrained and $K\ge 1$ is fixed.
Then any minimizer $f_i^\star$ of Eq.~\eqref{eq:local_infonce} satisfies, for $\mu_i$-a.e.\ $x$ and $\rho_i^{\boldsymbol{\pi}}$-a.e.\ $y$,
\begin{equation}
    f_i^\star(x,y)=\log\frac{M_i^{\boldsymbol{\pi}}(y\mid x)}{\rho_i^{\boldsymbol{\pi}}(y)}+C(x),
    \label{eq:optimal_critic}
\end{equation}
for some function $C:\mathcal{S}_i\to\mathbb{R}$ depending only on $x$.
\end{proposition}

\begin{proof}
For a fixed $x$, Eq.~\eqref{eq:local_infonce} is the cross-entropy loss of a $(K{+}1)$-way classifier that must identify the unique sample drawn from $M_i^{\boldsymbol{\pi}}(\cdot\mid x)$ among $K$ distractors drawn from $\rho_i^{\boldsymbol{\pi}}$.
The Bayes-optimal logits are proportional to the likelihood ratio, yielding Eq.~\eqref{eq:optimal_critic} up to an $x$-only constant.
\end{proof}

Following \citet{myers2024learning}, parameterize $f_i(x,y)=c_{\eta_i}(y)-d_{\phi_i}(x,y)$, where $d_{\phi_i}$ is interpreted as a temporal distance. Combining Eq.~\eqref{eq:optimal_critic} gives:
\begin{equation}
    d_{\phi_i}(x,y)= -\log M_i^{\boldsymbol{\pi}}(y\mid x)
    + \bigl(c_{\eta_i}(y)+\log \rho_i^{\boldsymbol{\pi}}(y)\bigr) - C(x),
    \label{eq:d_recovers}
\end{equation}
so $d_{\phi_i}$ recovers $-\log M_i^{\boldsymbol{\pi}}(y\mid x)$ up to additive $x$-only and $y$-only terms. 
Since $M_i^{\boldsymbol{\pi}}$ is defined by the induced marginal dynamics (Eq.~\eqref{eq:marginal_transition}), $d_{\phi_i}$ captures global dependencies mediated through $\boldsymbol{\pi}$.
\section{Benchmark Environment Details}\label{sec:appendix_B}
\subsection{Normalized Average Team Return}
For each task, we report the normalized average team return computed via task-wise linear normalization~\cite{fu2020d4rl, bettini2024benchmarl}:
\begin{equation}
    \tilde{R}=\frac{R-R_\text{rand}}{R_\text{ref}-R_\text{rand}}.
\end{equation}
Here, $R$ denotes the episodic team return (sum of environment rewards over an episode), $R_\text{rand}$ is the performance of a joint random policy, and $R_\text{ref}$ is a fixed task-specific reference performance.
Table~\ref{tab:total_steps_and_return_norm} reports the episode horizon (maximum timesteps before termination) and the normalization baselines ($R_\text{rand}, R_\text{ref}$) used to compute $\tilde{R}$ for each task.

\begin{table}[h]
    \centering
    \caption{Episode horizon and task-specific normalization baselines.}
    \begin{tabular}{lcc}
        \toprule
        \textbf{Task} & \textbf{Episode Horizon} & [$R_\text{rand}, R_\text{ref}$] \\
        \midrule
        MPE-corridor    & 256   & [-512, 0] \\
        MPE-tag         & 256   & [-768, 640] \\
        VMAS-balance    & 250   & [0, 130] \\
        VMAS-give way   & 600   & [0, 400] \\
        VMAS-wheel      & 600   & [0, 500] \\
        VMAS-transport  & 600   & [0, 1500] \\
        SMAX-\textit{3s5z\_vs\_3s6z} & 128 & [0, 1.5] \\
        SMAX-\textit{5m\_vs\_6m} & 128 & [0, 1.8] \\
        SMAX-\textit{10m\_vs\_11m} & 128 & [0, 2] \\
        SMAX-\textit{smacv2\_10\_units} & 128 & [0, 1.5] \\
        MABrax-ant\_4x2 & 300 & [0, 3300] \\
        MABrax-halfcheetah\_6x1 & 300 & [0, 3100] \\
        \bottomrule
    \end{tabular}
    \label{tab:total_steps_and_return_norm}
\end{table}

\subsection{Multi Particle Environments (MPE)~\cite{lowe2017multi}}
\begin{itemize}
    \item \textbf{MPE-corridor}: 
    This environment is an extension of the \textit{simple spread} task to evaluate multi-agent coordination under severe spatial constraints. 
    Eight agents are split into two groups and initialized at opposite vertical ends of the arena ($y=2.0$ and $y=-2.0$). 
    Goal locations follow mirror symmetry.
    Each agent's goal is the mirrored spawn of an agent from the opposite side, forming a point-symmetric crossing pattern.
    The map contains a central corridor of width $0.8$.
    At most two agents can pass concurrently, requiring temporal coordination to avoid deadlocks.

    We use a dense progress reward equal to the per-step reduction in distance to each agent's goal.
    We add penalties of $-1.5$ for inter-agent collisions and $-0.1$ for boundary violations.
    When an agent reaches its goal, it receives an additional per-step reward of $+0.5$.
    Finally, we add a team bonus of $+2.0$ only when all eight agents have reached their goals, encouraging collective completion.
    
    \item \textbf{MPE-tag}: 
    We utilize a fully cooperative predator-prey scenario based on FACMAC~\cite{peng2021facmac}. 
    The environment contains six predators (learning agents), two prey (scripted agents), and three obstacles. 
    Each prey samples $100$ feasible candidate next positions in a unit disk and moves toward the one that maximizes its distance to the nearest predator.

    Predators share a team reward composed of a sparse capture bonus and a dense shaping term.
    Specifically, we add a bonus of $+10$ when any prey is caught.
    We also include a per-step shaping reward $-d_{\min}$, where $d_{\min}$ is the minimum predator--prey distance.
\end{itemize}

\begin{figure*}[t]
    \centering
    \subfigure[\textit{Balance}.]
    {\includegraphics[width=0.2\textwidth]{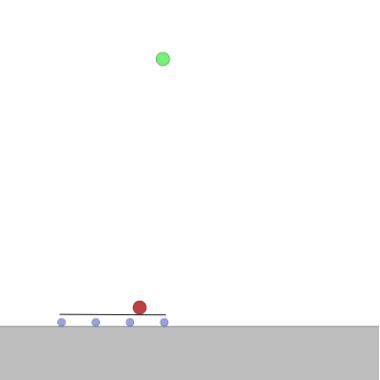}}
    \subfigure[\textit{Give way}.]
    {\includegraphics[width=0.2\textwidth]{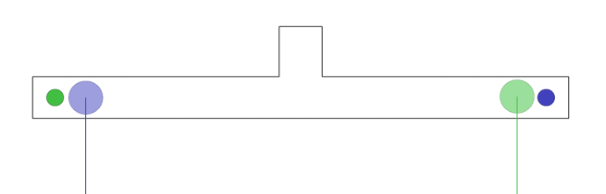}}
    \subfigure[\textit{Transport}.]
    {\includegraphics[width=0.2\textwidth]{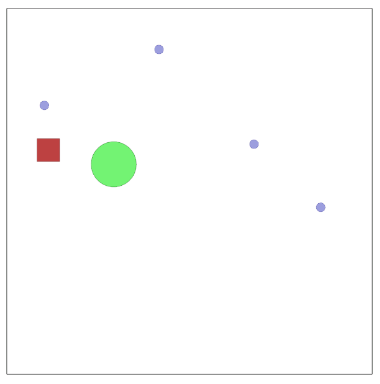}}
    \subfigure[\textit{Wheel}.]
    {\includegraphics[width=0.2\textwidth]{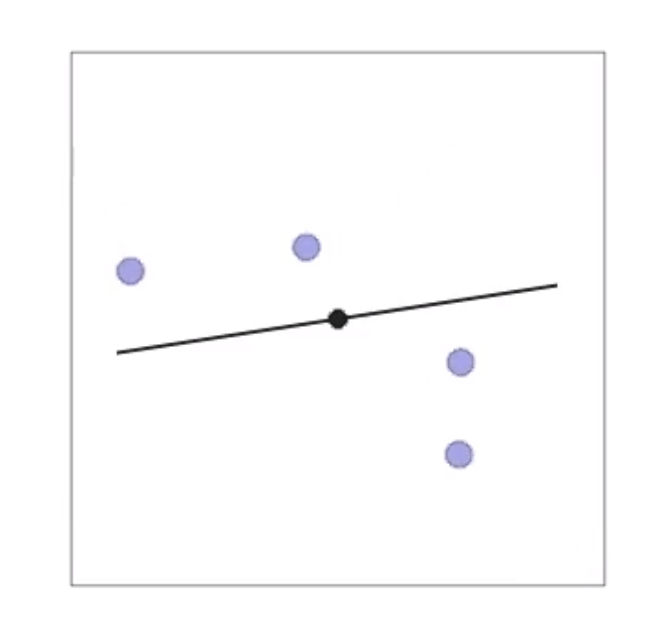}}
    \caption{Multi-agent tasks from the VMAS simulator used in our experiments.}
    \label{fig:VMAS_tasks}
\end{figure*}

\subsection{Vectorized Multi-Agent Environments (VMAS)~\cite{bettini2022vmas}}

\begin{itemize}
    \item \textbf{Transport}: 
    Four agents must jointly push a single heavy package ($m=50$) to a goal position. 
    The team reward is the per-step reduction in the package--goal distance, scaled by $20$ for comparability across benchmarks. 
    Since a single agent cannot move the package effectively, success requires coordinated force application.
    
    \item \textbf{Wheel}: 
    Four agents must cooperatively spin an anchored rod ($m=30$) to match a target angular velocity $\omega = 0.05$. 
    The reward is calculated as $1 - ((\text{abs}(\dot{\theta}) - \omega) / \omega)^2$, yielding a maximum value of 1.0 at the target velocity. 
    To prevent agents from drifting away and to stabilize contact, we add a surrounding boundary with a shared collision penalty $-0.1$.
    
    \item \textbf{Balance}: 
    Operating under vertical gravity, four agents support a movable rod ($l=0.8$) and carry a spherical package ($m=5$) to a high-altitude goal position. 
    The reward is the per-step reduction in the package--goal distance.
    We apply a terminal penalty of $-10$ if either the rod or the package contacts the floor, enforcing balance throughout the transport.
    
    \item \textbf{Give way}: 
    Two agents start at opposite ends and must swap positions (each agent's goal is the other's start).
    A central bottleneck admits only one agent at a time, requiring explicit yielding to avoid deadlock.
    The team reward is twice the per-step reduction in the agent--goal distance.
    We add a $-0.01$ penalty for collisions with agents or walls, and give an additional per-step reward of $+1.0$ to agents while they remain at their goals.
    Because solving the task requires asymmetric roles, we disable parameter sharing for this environment.
\end{itemize}

\subsection{StarCraft Multi-Agent Challenge-like JAX environment (SMAX) ~\cite{rutherford2024jaxmarl}}
SMAX is a JAX-native implementation of the StarCraft Multi-Agent Challenge (SMAC)~\cite{samvelyan2019starcraft, ellis2023smacv2}, designed for high-throughput multi-agent reinforcement learning. 
Unlike the original StarCraft II-based environment, SMAX eliminates the overhead of the game engine while retaining the core strategic elements of unit micromanagement.
We evaluate our method on four scenarios with asymmetrical team compositions across the main text and appendix:
\begin{itemize}
    \item \textbf{SMAX-\textit{3s5z\_vs\_3s6z}}: A heterogeneous combat task with Stalkers and Zealots.
    \item \textbf{SMAX-\textit{5m\_vs\_6m}} \& \textbf{SMAX-\textit{10m\_vs\_11m}}: 
    Homogeneous Marine battles against a numerically superior opponent. Success depends on ``focus fire'' tactics and kiting to overcome the numerical disadvantage.
    \item \textbf{SMAX-\textit{smacv2\_10\_units}}: This task utilizes the procedural generation features of SMACv2, where unit types and start positions are randomized, requiring the policy to generalize across diverse combat configurations.
\end{itemize}

\subsection{Multi-Agent Brax (MABrax)~\cite{rutherford2024jaxmarl}}
MABrax is a JAX-accelerated benchmark for continuous multi-agent robotic control, built upon the Brax physics engine~\cite{freeman2021brax}. 
Serving as a high-throughput alternative to Multi-Agent MuJoCo~\cite{peng2021facmac}, MABrax decomposes a single complex robotic morphology into multiple sub-components, each controlled by an independent agent. 
We evaluate \textsc{EFXPLORER} on two representative morphologies:
\begin{itemize}
    \item \textbf{MABrax-ant\_4x2}: A quadrupedal ``Ant'' robot is partitioned among 4 agents. Each agent is responsible for controlling 2 joints corresponding to a single leg.
    \item \textbf{MABrax-halfcheetah\_6x1}: A planar bipedal ``HalfCheetah'' robot is divided among 6 agents, with each agent controlling exactly 1 joint.
\end{itemize}
\section{Differences from Prior Epigraph-based Constrained RL}\label{sec:appendix_C}
\begin{figure}[hbt!]
    \begin{center}
        \centerline{\includegraphics[width=1.0\linewidth]{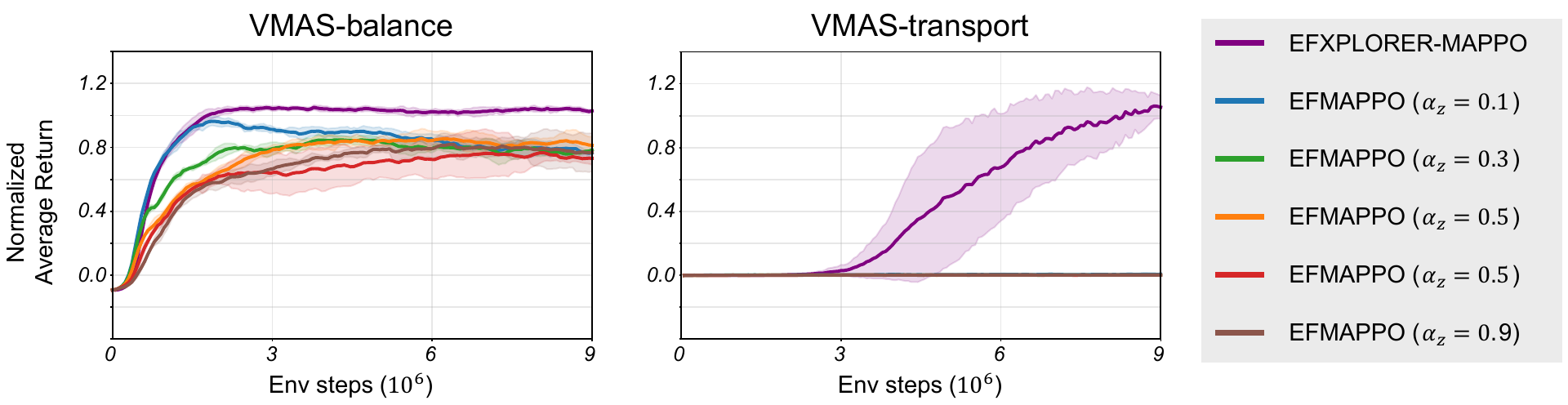}}
    \end{center}
    \vskip -0.3in
    \caption{Performance comparison with prior epigraph coupling mechanisms across varying budget update rates ($\alpha_z$). Compared to our local surrogate (\textsc{EFXPLORER-MAPPO}), the safety-oriented dynamic programming approach (EFMAPPO) severely restricts policy updates, failing to induce necessary exploration on \textit{VMAS-transport} and showing no improvement on \textit{VMAS-balance}.}
    \label{fig:appendix_DefMARL_ablation}
\end{figure}	

Epigraph reformulation has been used in constrained RL to avoid manual mixing coefficients by introducing an auxiliary budget and coupling feasibility with optimization via a conservative bottleneck (often $\min$)~\cite{so2023solving, zhang2025solving}.
Most prior work targets \emph{safety/cost}-type constraints defined by an instantaneous signal $h(s_t,a_t)$, and enforces feasibility through trajectory-level aggregation (e.g., a worst-case margin over time).
Our setting and epigraph construction differ from prior epigraph constraints in the following ways:
\begin{itemize}
    \item \textbf{Constraint Type.} We impose a \emph{return-level non-degradation} constraint,
    $J_{\mathrm{ext}}(\boldsymbol{\pi}) \ge J_{\mathrm{ext}}(\hat{\boldsymbol{\pi}})$,
    rather than a trajectory-wide constraint derived from an instantaneous cost/safety signal.
    \item \textbf{Step-wise Feasibility Signal.} We use the reference advantage
    $A_{\hat{\boldsymbol{\pi}}}^{\mathrm{ext}}(s_t,\mathbf{a}_t)$ as a stepwise certificate (via the performance-difference identity),
    which is inherently horizon-aware through $Q_{\hat{\boldsymbol{\pi}}}^{\mathrm{ext}}$ and $V_{\hat{\boldsymbol{\pi}}}^{\mathrm{ext}}$.
    \item \textbf{Exploration Coupling.} Our bottleneck directly couples task progress and exploration surplus:
    $\mathcal{B}_i^{\mathrm{epi}}=\min\{A_{\hat{\boldsymbol{\pi}}}^{\mathrm{ext}},\, S_i^{\mathrm{int}}\}$ (Eq.~\eqref{eq:epigraph_bottleneck}),
    yielding an exploration update that is \emph{gated} by feasibility rather than tuned by a coefficient.
    \item \textbf{Value Propagation (Dynamic Programming).} Prior methods~\cite{so2023solving, zhang2025solving} redefine the Bellman operator to strictly propagate a worst-case safety margin backward through time. In contrast, we retain standard value functions ($V^\text{ext}, V^\text{int}$) and apply the epigraph bottleneck locally via advantages, avoiding overly pessimistic value estimates.
\end{itemize}

\paragraph{Empirical Comparison with Safety-oriented DP.}
To validate our localized surrogate, we construct a baseline termed \textbf{EFMAPPO}. 
We adapt the multi-agent extension of EFPPO algorithm~\cite{so2023solving} rather than Def-MARL~\cite{zhang2025solving}, as Def-MARL define local per-agent function in their constraint and is thus incompatible with our shared extrinsic advantage formulation.
For a fair comparison, EFMAPPO employs our exact objective and budget adaptation ($\alpha_z$), but replaces our local advantage bottleneck with EFPPO's safety-oriented dynamic programming (DP), which strictly propagates the minimum feasibility margin over trajectories.

As shown in Fig.~\ref{fig:appendix_DefMARL_ablation}, safety-oriented DP is ill-suited for exploration.
EFMAPPO performs similarly to or worse than MAPPO on \textit{VMAS-balance} and completely flatlines on the exploration-heavy \textit{VMAS-transport} across all $\alpha_z$ values. 
These failures occur because EFPPO's Bellman operator treats temporary drops in task advantage as catastrophic constraint violations. 
Consequently, agents either rigidly ignore intrinsic bonuses to maintain stability (\textit{balance}) or suffer vanishing gradients when exploration is required (\textit{transport}). 
In contrast, \textsc{EFXPLORER} successfully navigates this trade-off via its local advantage bottleneck.
\section{Further Analysis of Alternative Optimization Methods}\label{sec:appendix_D}
We provide an analytical comparison between our epigraph formulation (\textsc{EFXPLORER}) and the Lagrangian relaxation used in \textbf{MAPPO-Lagrangian}. 
Recall the constrained exploration objective from Eq.~\eqref{eq:constrained_optimization}, which enforces non-degradation of team performance:

\begin{equation}
\max_{\boldsymbol{\pi}\in\Pi} \sum_{i\in\mathcal{N}} J_{i,\text{int}}(\pi^i)\quad \text{s.t.} \quad J_\text{ext}(\boldsymbol{\pi}) \geq J_\text{ext}(\hat{\boldsymbol{\pi}}).
\end{equation}

\paragraph{Lagrangian Relaxation and Multiplier Sensitivity.}
A standard primal--dual relaxation yields:
\begin{equation}
    \min_{\lambda \ge 0} \max_{\boldsymbol{\pi} \in \Pi} \mathcal{L}(\boldsymbol{\pi}, \lambda) \triangleq \sum_{i\in\mathcal{N}}J_{i,\mathrm{int}}(\pi^i) + \lambda \Delta J_{\mathrm{ext}}(\boldsymbol{\pi}).
    \label{eq:dual_form_app_revised}
\end{equation}
Leveraging the stepwise certificate from Theorem~\ref{thm:suff_stepwise}, the dual formulation can be equivalently expressed for an individual agent through an advantage-based objective. 
To align with \textsc{EFXPLORER}, we consider the per-agent surrogate $A_{i}^{\mathrm{int}} + \lambda A^{\mathrm{ext}}$, where the dual variable $\lambda$ is adjusted via dual gradient descent:
\begin{equation}
    \lambda \leftarrow \Big[\lambda - \eta_\lambda v(\boldsymbol{\pi})\Big]_+,
\end{equation}
using the mean advantage as the violation signal $v(\boldsymbol{\pi}) \triangleq \mathbb{E}_{(s_t,a_t)\sim\tau}[A^{ext}(s_t,a_t)]$.

This framework presents two challenges for coordination in multi-agent environments:
\begin{itemize}
    \item \textbf{Multiplier Sensitivity:} 
    The policy gradient is sensitive to the magnitude of $\lambda$, often requiring a conservative dual step size to avoid unstable updates~\cite{so2023solving, zhang2025solving}. 
    This introduces a trade-off: small $\eta_\lambda$ stabilizes learning but slows multiplier adaptation, delaying responses to performance degradation.
    \item \textbf{Batch-wide Averaging:} Since $\lambda$ is updated from batch-averaged violation signals~\cite{li2021augmented}, it provides a batch-level control signal.
    As a result, it cannot gate exploration at individual timesteps, which is critical for coordination in cooperative MARL.
\end{itemize}

\paragraph{Key Distinction in \textsc{EFXPLORER}.}
In contrast, \textsc{EFXPLORER} introduces the epigraph-based \textit{min-bottleneck} mechanism:
\begin{equation}
    \mathcal{B}_i^{\mathrm{epi}} = \min \{ A_{\hat{\pi}}^{\mathrm{ext}}, S_i^{\mathrm{int}} \}.
\end{equation}
By performing this comparison at each timestep, we enable instantaneous gating. 
This yields immediate suppression of exploration when task progress is threatened, avoiding the instability and tuning sensitivity induced by the dual variable~$\lambda$.
Further comparisons across different $\lambda_0$ values are provided in Appendix~\ref{sec:appendix_F}.
\section{Implementation Details}\label{sec:appendix_E}
\subsection{Pseudocode of \textsc{EFXPLORER}}
To clarify our alternating maximization scheme, we outline the training procedure of \textsc{EFXPLORER}-MAPPO in two parts. 
Algorithm~\ref{alg:efxplorer_inner} details the \textbf{Inner Optimization}, where the policy and value networks are updated using the epigraph-gated advantage. Algorithm~\ref{alg:efxplorer_outer} details the \textbf{Outer Optimization}, where the exploration budgets for each agent are adaptively expanded based on task feasibility. 
During training, the framework alternates between data collection and these two optimization loops.

\begin{algorithm}[t] 
\captionsetup{font=normal}
\caption{\textsc{EFXPLORER}-MAPPO: Inner Optimization}
\label{alg:efxplorer_inner}
{\setlength{\baselineskip}{0.93\baselineskip}
\begin{algorithmic}
\STATE \textbf{Init:} actor $\pi$, critics $V^{\text{ext}}, \{V^{\text{int}}_i\}_{i\in\mathcal{N}}$, successor distance~$\{d_{\phi_i}\}_{i\in\mathcal{N}}$, \\ \hspace*{2em}replay buffer $\mathcal{R}$, budgets $\{z^i\}_{i\in\mathcal{N}}$
\WHILE{not converged}
\STATE $\hat{\boldsymbol{\pi}} \leftarrow \boldsymbol{\pi}$ \hfill \textcolor{gray}{\# reference joint policy}
\STATE Sample rollouts $\mathcal{D}=\{\tau_m\}_{m=1}^{M} \sim \boldsymbol{\pi}$; \quad $\mathcal{R}\leftarrow \mathcal{R}\cup\mathcal{D}$
\STATE Update $d_{\phi_i}$ on $\mathcal{R}$

\FOR{each rollout $\tau_m\in\mathcal{D}$}
\STATE Compute $\{r^{\text{int}}_{i,t}\}$ (Eq.~\eqref{eq:indiv_novelty}) and propagate budgets:\\
$z^{i,(m)}_{0}=z^i, \quad \gamma z_{t+1}^{i,(m)} = z_{t}^{i,(m)} - r^{\text{int}}_{i,t} \quad \forall i,t$
\STATE Compute $\{A^{\text{epi-GAE}}_{i,t}\}$ using Epigraph-GAE \hfill (Eq.~\eqref{eq:final_epi_adv})
\ENDFOR
\STATE Update critics $V^{\text{ext}},\{V^{\text{int}}_i\}$ by regression to their returns
\STATE Perform $K$-epoch MAPPO updates on $\pi$ using $\{A^{\text{epi-GAE}}_{i,t}\}$.
\STATE Update budgets $\{z^i\}$ by Algorithm~\ref{alg:efxplorer_outer}. \hfill \textcolor{gray}{\# outer loop}
\ENDWHILE
\end{algorithmic}
}
\end{algorithm}

\begin{algorithm}[t] 
\captionsetup{font=normal}
\caption{\textsc{EFXPLORER}: Outer Optimization}
\label{alg:efxplorer_outer}
{\setlength{\baselineskip}{1.105\baselineskip}
\begin{algorithmic}
\STATE \textbf{Input:} budgets $\{z^i\}$, reference level $\bar{J}^\text{ext}_k$ \quad \textbf{Init:} $\mathcal{F}_k \leftarrow \emptyset$
\FOR{each rollout $\tau_m\in\mathcal{D}$}
    \STATE \textbf{1. Estimate candidate budgets from intrinsic returns:}\\
    $\hat{z}^{i,k,m}_{0}\leftarrow \sum_{t=0}^{T_m-1}\gamma^t r^{\text{int}}_{i,t}\quad \forall i\in\mathcal{N}$ 
    \STATE \textbf{2. Check feasibility from extrinsic returns}:\\
    \STATE Compute task return: $\hat{J}^{\text{ext}}_{k,m} \leftarrow \sum_{t=0}^{T_m-1} \gamma^tr^{\text{ext}}_{t}$
    \STATE \textbf{if} $\hat{J}^{\text{ext}}_{k,m} > \bar{J}^\text{ext}_k$ \textbf{then} $\mathcal{F}_k \leftarrow \mathcal{F}_k \cup \{m\}$
\ENDFOR
\STATE \textbf{3. Budget update:}
\IF{$\mathcal{F}_k\neq\emptyset$}
    \STATE $m^\ast\in \arg\max_{m\in\mathcal{F}_k}\sum_{i\in\mathcal{N}}\hat{z}^{i,k,m}_{0}$ \hfill (Eq.~\eqref{eq:select_best_rollout})
    \STATE $z^{i,k+1}\leftarrow \alpha_z\cdot \hat{z}^{i,k,m^\ast}_{0}+(1-\alpha_z)\cdot z^{i,k}$ \quad $\forall i$ \hfill (Eq.~\eqref{eq:init_z_ema})
    \STATE Update reference $\bar{J}^\text{ext}_{k+1}$ using $\hat{J}^{\text{ext}}_{k,m^\ast}$ (e.g., moving average)
\ENDIF
\end{algorithmic}
}
\end{algorithm}

\subsection{Computation Resources}
All models were trained using vectorized frameworks to leverage hardware acceleration. 
The experiments were conducted on a workstation equipped with dual 32-core AMD EPYC 9354 processors and $8 \times$ NVIDIA L40S GPUs (48GB VRAM each). 
Under this configuration, \textsc{EFXPLORER}-MAPPO required 50 minutes of training on VMAS and 2 hours on MPE environments and 1 hours on SMAX environments, approximately 1.7$\times$ the training time of standard MAPPO. 
Due to the additional buffers used for intrinsic reward computation, its VRAM usage was roughly twice that of MAPPO, reaching about 10GB in our experiments, while the inference cost remained unchanged.

\subsection{Hyperparameters for \textsc{EFXPLORER}}
Table~\ref{tab:param-backbone} summarizes the hyperparameters used to train the backbone algorithm, which are shared across \textsc{EFXPLORER} and all baseline methods.
Table~\ref{tab:param-efxplorer-mappo} and Table~\ref{tab:param-efxplorer-ippo} report the \textsc{EFXPLORER}-specific hyperparameters when the backbone is MAPPO and IPPO, respectively.

\begin{table}[h]
    \centering
    \caption{Hyperparameters for training backbones.}
    \label{tab:param-backbone}
    \begin{tabular}{lccc}
        \toprule
        \textbf{Parameter}    & \textbf{MPE} & \textbf{VMAS} & \textbf{SMAX} \\
        \midrule
        discount factor $\gamma$  & 0.99 & 0.99 & 0.99\\
        GAE coefficient $\lambda_{\text{GAE}}$ & 0.95 & 0.9 & 0.95\\
        gradient clip norm        & 0.6 & 10 & 0.6 \\
        epsilon clip                & 0.3 &  0.15 & 0.1\\
        coefficient of entropy      & 0.015 & - & 0.01\\
        extrinsic value loss coefficient &0.5 & - & 0.5\\
        RNN hidden state dimension size & 256 & - & 128\\
        FC layer dimension size     & 64 & 512 & 128\\
        number of GRU layers  &1 & - & 1\\
        number of  FC layers  &2 &  2 & 2\\
        number of total timesteps       &$3 \times 10^7$ & $9 \times 10^6$ & $3 \times 10^7$ \\
        number of epoch $K$       & 10 & 45 & 10\\
        number of minibatch       & 4 & 4000 & 4\\
        number of parallel envs $M$& 200 & 100 & 64\\
        value loss                  & clipped MSE loss & clipped MSE loss & clipped MSE loss\\ 
        optimizer                   & Adam   &  Adam & Adam\\
        optimizer epsilon           & $10^{-5}$  &  $10^{-6}$ & $10^{-5}$\\
        \bottomrule
    \end{tabular}
\end{table}

\begin{table}[h]
    \centering
    \caption{Hyperparameters for training \textsc{EFXPLORER}-MAPPO.}
    \label{tab:param-efxplorer-mappo}
    \begin{tabular}{lccc}
        \toprule
        \textbf{Parameter}  & \textbf{MPE} & \textbf{VMAS}   & \textbf{SMAX}     \\
        \midrule
        \textbf{Successor distance network $d_\phi$} & & \\
        latent dimension size  & 64 & 128 & 256 \\
        batch size $B$             & 512   &  512 & 512\\
        learning rate          & $1\times 10^{-2}$  & $2\times10^{-3}$ & $5\times10^{-3}$ \\
        number of epoch            & 25   & 16 & 10 \\
        number of residual blocks  & 10   & 10 & 4\\
        \textbf{Epigraph-related parameters} & & \\
        $\alpha_z$ & 0.7 &\begin{tabular}{@{}c@{}} 0.1 (balance, wheel) \\ 0.4 (transport, give way) \end{tabular}& 0.3\\
        intrinsic value learning rate & $3\times10^{-3}$ & \begin{tabular}{@{}c@{}} $5\times10^{-5}$  \\ $1\times10^{-4}$ (give way) \end{tabular} & $5\times10^{-4}$\\
        actor learning rate               & 
        \begin{tabular}{@{}c@{}} 0.001 \\ (0.0005 for tag) \end{tabular} & $1\times10^{-4}$ & $1.5\times10^{-4}$\\
        value learning rate               & \begin{tabular}{@{}c@{}} 0.001 \\ (0.0005 for tag) \end{tabular}& $4\times10^{-4}$ & $4\times10^{-4}$\\
        \bottomrule
    \end{tabular}
\end{table}

\begin{table}[h]
    \centering
    \caption{Hyperparameters for training \textsc{EFXPLORER}-IPPO.}
    \label{tab:param-efxplorer-ippo}
    \begin{tabular}{lccc}
        \toprule
        \textbf{Parameter}  & \textbf{MPE} & \textbf{VMAS}    & \textbf{SMAX}    \\
        \midrule
        \textbf{Successor distance network $d_\phi$} & & \\
        latent dimension size  & 64 & 128 & 126 \\
        batch size $B$             & 512   &  512 & 512\\
        learning rate          & $1\times 10^{-2}$  & $2\times10^{-3}$ & $5\times10^{-3}$ \\
        number of epoch            & 25   & 16 & 10\\
        number of residual blocks  & 10   & 10 & 4\\
        \textbf{Epigraph-related parameters} & & \\
        $\alpha_z$ & 0.7 & 0.1 & 0.1\\
        intrinsic value learning rate & 0.003 & $5\times10^{-5}$ & $5\times10^{-4}$\\
        actor learning rate               & 0.001 (0.0005 for tag) & $1\times10^{-4}$ & $1.5\times10^{-4}$\\
        value learning rate               & 0.001 (0.0005 for tag) & $1\times10^{-4}$ & $4.5\times10^{-3}$\\
        \bottomrule
    \end{tabular}
\end{table}

\subsection{Implementation Details of Baselines}
We implement all baselines following their original papers and, when available, official codebases.
To ensure a fair comparison, all methods share the same backbone training pipeline and the hyperparameters in Table~\ref{tab:param-backbone}.
For completeness, we provide a dedicated hyperparameter table for each baseline; any deviation from the shared settings is explicitly noted in the corresponding table.
We next summarize baseline-specific implementation details.
\begin{itemize}
    \item \textbf{COIN}~\cite{li2023two}: We follow the original implementation described in~\cite{li2023two}. The hyperparameter configuration is reported in Table~\ref{tab:param-coin}.
    \item \textbf{HASAC}~\cite{liu2023maximum}:  We implement HASAC according to~\cite{liu2023maximum}. We reduced the size of the replay buffer compared to the original repository to reduce the variance of the update. The hyperparameter configuration is reported in Table~\ref{tab:param-hasac}. 
    \item \textbf{DiCO}~\cite{bettini2024controlling}: DiCO controls behavioral diversity by leveraging a measure known as System Neural Diversity (SND)~\cite{bettini2025system}. 
    To ensure a fair comparison, we performed a grid search for the SND coefficient over the set $\{0, 0.1, 0.2, 0.5, 1.0, 2.0, 5.0, 10.0\}$. 
    We selected the hyperparameters that yielded the best average performance across 5 independent random seeds. 
    The selected task-specific coefficients are reported in Table~\ref{tab:param-dico}, along with the remaining hyperparameters.
    \item \textbf{MAPPO-Lagrangian}~\cite{gu2023safe}: We implement MAPPO-Lagrangian as stated in Appendix~\ref{sec:appendix_D}. For fair comparison, we utilized the same learning parameters with EFXPLORER-MAPPO, and use $\eta_\lambda=10^{-3}$.
\end{itemize}

\begin{table}[h]
    \centering
    \caption{Hyperparameters for training COIN.}
    \label{tab:param-coin}
    \begin{tabular}{lccc}
        \toprule
        \textbf{Parameter}    & \textbf{MPE} & \textbf{VMAS} & \textbf{SMAX} \\
        \midrule
        influence hidden dimension size     & 64  &  64 & 128 \\
        influence learning rate             & $1 \times 10^{-3}$ &  $1 \times 10^{-2}$ & $5 \times 10^{-3}$\\
        curiosity hidden dimension size     & 64 & 64 & 128\\
        curiosity learning rate             & $1 \times 10^{-3}$ & $1 \times 10^{-2}$ & $5 \times 10^{-3}$\\
        influence scale                     & 0.0005 (0.01 for tag) & 0.0005 & 0.0005\\
        curiosity scale                     & 0.0005 (0.01 for tag) & 0.0005 & 0.0005\\
        warmup rollouts                     & 5 & 0 & 5\\
        network update frequeny             & 1 & 200 & 1\\
        gradient clip norm                  & 0.5 & 10 & 0.5\\
        critic learning rate                & 0.0003 & 0.0005 & 0.0005\\
        \bottomrule
    \end{tabular}
\end{table}

\begin{table}[h]
    \centering
    \caption{Hyperparameters for training HASAC.}
    \label{tab:param-hasac}
    \begin{tabular}{lccc}
        \toprule
        \textbf{Parameter}    & \textbf{MPE} & \textbf{VMAS} & \textbf{SMAX} \\
        \midrule
        number of parallel environments $M$     & 32 (16 for tag) & 60 & 64\\
        number of epoch $K$                & 6 & 1000 & 300\\
        RNN hidden state dim & 256 (64 for tag) & - & 128\\
        initial entropy temperature         & 0.1 & $1\times10^{-3}$ & 0.1\\
        target entropy ratio                & 0.5 & 0.2 & 0.05\\
        value loss                          & \text{plain MSE loss} & smooth $\ell1$ & \text{plain MSE loss}\\
        actor learning rate                 & 0.001 (0.0005 for tag) & 0.0005 & 0.0005\\
        critic learning rate                & 0.0005 (0.0003 for tag) & 0.0005 & 0.0004\\
        buffer size                         & 50000 & 500000 & 40000\\
        buffer batch size                   & 256 & 768 & 256\\
        learning starts                     & 10000 & 0 & 50000\\
        \bottomrule
    \end{tabular}
\end{table}

\begin{table}[h]
    \centering
    \caption{Hyperparameters for training DiCO.}
    \label{tab:param-dico}
    \begin{tabular}{lccc}
        \toprule
        \textbf{Parameter}    & \textbf{MPE} & \textbf{VMAS} & \textbf{SMAX}\\
        \midrule
        desired SND                         & 0.1 & \begin{tabular}{@{}c@{}} 0.1 (balance, transport) \\ 1.0 (wheel, give way) \end{tabular}  & 0.1\\
        influence learning rate             & $1 \times 10^{-3}$ & $5 \times 10^{-5}$ & $1 \times 10^{-3}$ \\
        learning rate                       & 0.0005 & 0.0005 & 0.0005\\
        bootstrap from desired SND      & \text{true} & \text{true} & \text{true}\\
        \bottomrule
    \end{tabular}
\end{table}
\section{Additional Experimental Results}\label{sec:appendix_F}
In this section, we provide extended empirical analyses to further validate the efficacy, robustness, and generalizability of \textsc{EFXPLORER}. 
First, we evaluate our method on additional continuous and discrete benchmarks (MABrax and SMAX).
Second, we extend key ablations to additional environments and include a softmin ablation.
Finally, we provide a rigorous hyperparameter sensitivity analysis to demonstrate our framework's robustness compared to standard heuristic approaches.

\subsection{Additional Benchmark Evaluations}
\begin{figure}[hbt!]
    \begin{center}
        \centerline{\includegraphics[width=1.0\linewidth]{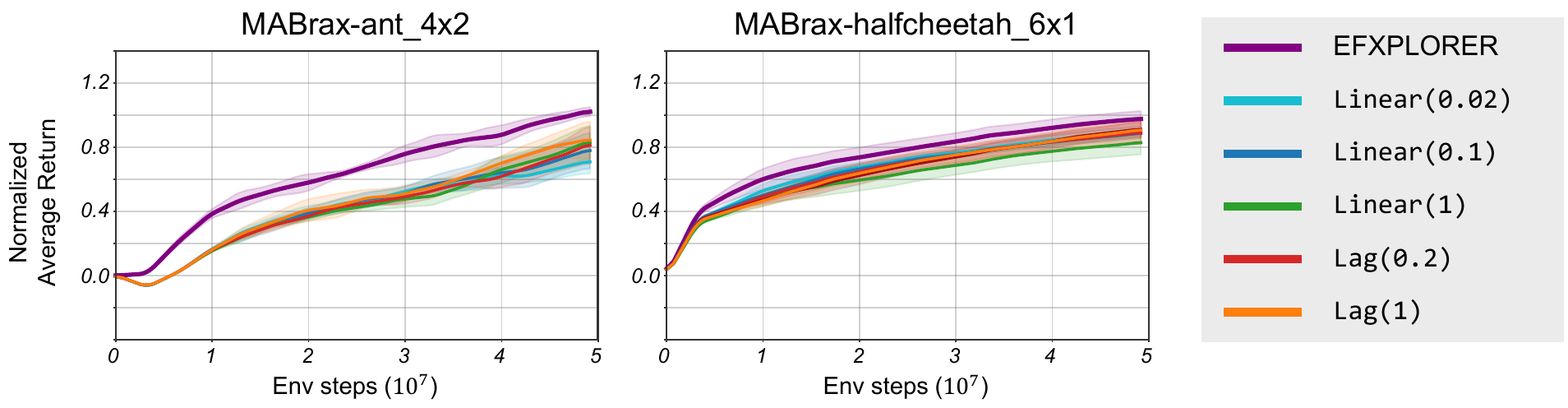}}
    \end{center}
    \vskip -0.3in
    \caption{Performance on continuous control tasks in MABrax. Solid lines show the mean over 5 random seeds, while shaded regions correspond to the standard deviation.}
    \label{fig:appendix_mabrax}
\end{figure}	

\paragraph{Continuous Control: MABrax.}
We evaluate \textsc{EFXPLORER} on continuous tight-coordination tasks (MABrax-ant\_4x2, MABrax-halfcheetah\_6x1).
In these environments, erratic exploration easily causes the shared robot body to fall (Fig.~\ref{fig:appendix_mabrax}). 
While heuristic baselines like MAPPO-Linear ($\beta$) and MAPPO-Lagrangian ($\lambda_0$) can achieve reasonable progress, they remain vulnerable to coordination collapse. 
By gating exploration via the epigraph bottleneck, \textsc{EFXPLORER} ensures stable locomotion throughout training, performing on par with or marginally outperforming the baselines.

\begin{figure}[hbt!]
    \begin{center}
        \centerline{\includegraphics[width=0.8\linewidth]{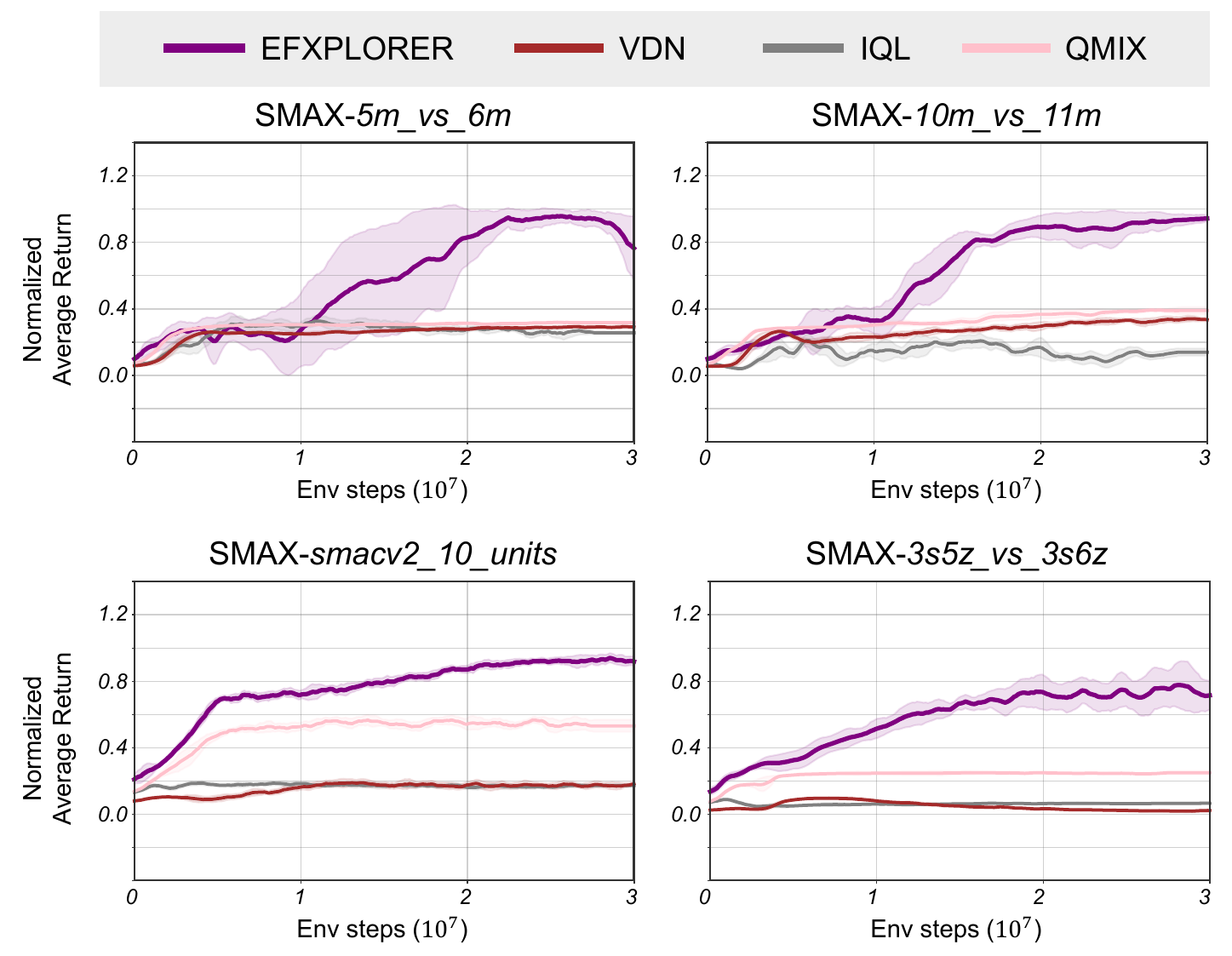}}
    \end{center}
    \vskip -0.3in
    \caption{Performance on additional SMAX micromanagement scenarios. Solid lines show the mean over 3 random seeds, while shaded regions correspond to the standard deviation.}
    \label{fig:appendix_smax_additional}
    \vspace{-10pt}
\end{figure}	

\paragraph{Discrete Micromanagement: SMAX.}
Fig.~\ref{fig:appendix_smax_additional} extends our evaluation to four challenging discrete micromanagement scenarios in SMAX. 
Compared to established MARL baselines (VDN~\cite{sunehag2017value}, IQL~\cite{tampuu2015multiagent}, QMIX~\cite{rashid2020monotonic}), \textsc{EFXPLORER} achieves competitive or superior performance. 
This confirms that our novelty-driven approach effectively discovers complex tactics without compromising team survival.

\subsection{Additional Ablation Studies}
\begin{figure}[hbt!]
    \begin{center}
        \centerline{\includegraphics[width=1.0\linewidth]{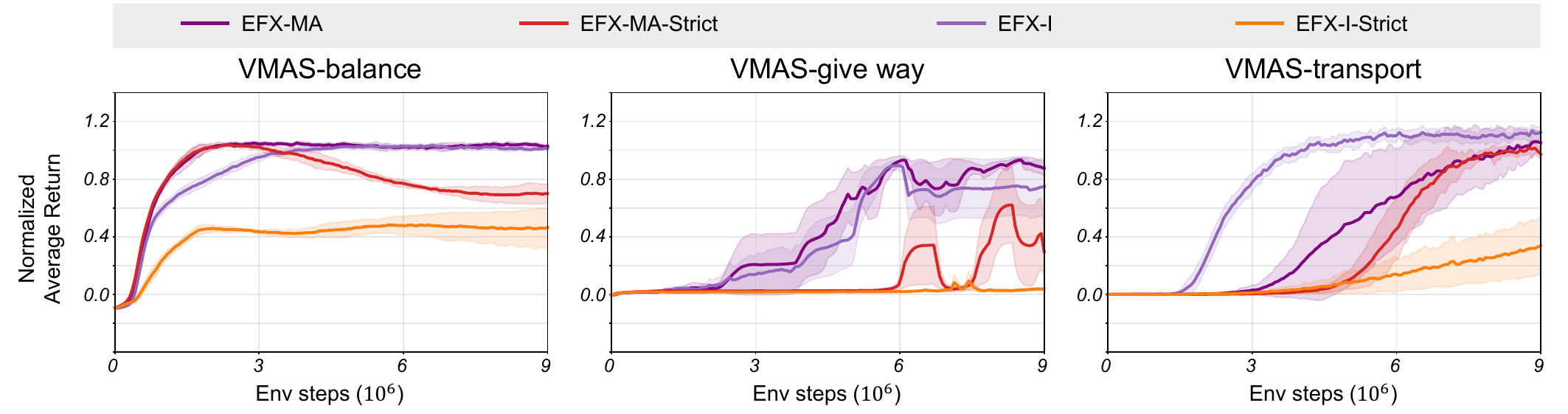}}
    \end{center}
    \vskip -0.3in
    \caption{Ablation of the reference performance metric ($\bar{J}^\text{ext}_k$) in the outer optimization loop. A strict best-return criterion is easily dominated by noisy estimates, heavily restricting budget growth. In contrast, our running reference provides smoother budget adaptation.}
    \label{fig:appendix_outer_additional}
\end{figure}	
\paragraph{Effect of the Running Reference.}
Fig.~\ref{fig:appendix_outer_additional} ablates the choice of the reference performance $\bar{J}^\text{ext}_k$ used to define the feasible set (Eq.~\eqref{eq:outer_feasible_set}). We compare our default running reference (e.g., a moving average) against a strict best-return criterion. Because empirical Monte Carlo returns are inherently noisy, tracking the strict absolute maximum establishes an overly pessimistic threshold. As shown in the results, this strict criterion severely restricts the growth of the exploration budget, slowing down exploration. Conversely, the running reference effectively smooths out noisy outliers, ensuring steady budget adaptation and continuous task progress.

\begin{figure}[hbt!]
    \begin{center}
        \centerline{\includegraphics[width=1.0\linewidth]{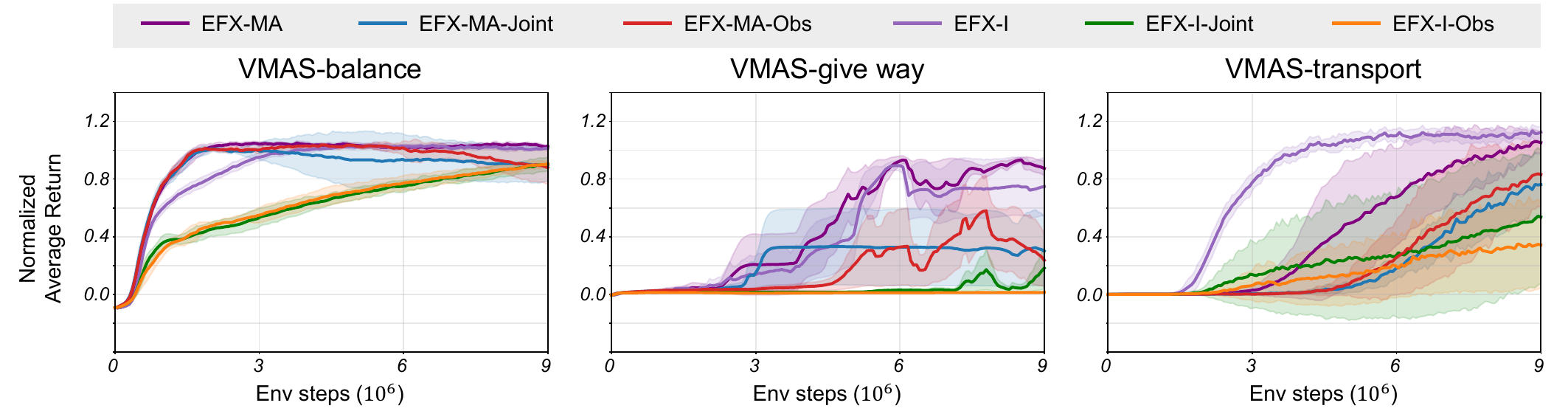}}
    \end{center}
    \vskip -0.3in
    \caption{Ablation of state factorization for successor features. Individual state spaces avoid the intractability of joint states and the non-stationarity of observation features.}
    \label{fig:appendix_state_fac_additional}
\end{figure}	
\paragraph{State Factorization.}
Fig.~\ref{fig:appendix_state_fac_additional} compares our agent-centric state factorization against Joint-state and Observation-level alternatives across three VMAS tasks. Joint-state estimation is intractable, while observation-level features inject teammate-induced non-stationarity that destabilizes learning. Individual states optimally balance scalability and stability.

\begin{figure}[hbt!]
    \begin{center}
        \centerline{\includegraphics[width=1.0\linewidth]{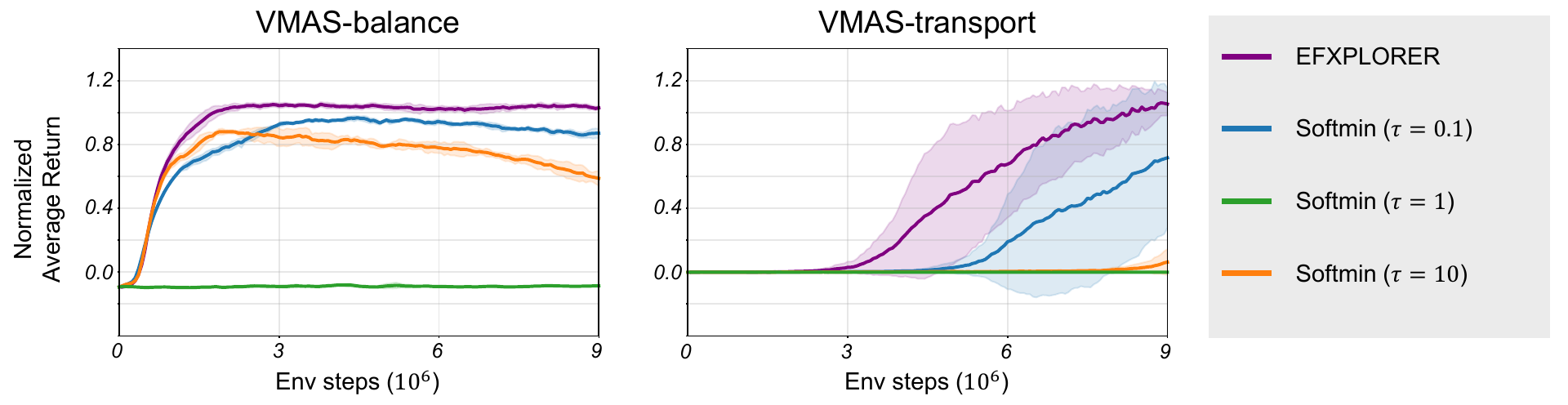}}
    \end{center}
    \vskip -0.3in
    \caption{Comparison of the hard $\min$ bottleneck against a \texttt{softmin} variant. The hard $\min$ eliminates the need for temperature tuning ($\tau$) and avoids both gradient mixing and coordination collapse.}
    \label{fig:appendix_softmin}
\end{figure}	
\paragraph{Hard \textit{min} vs. Softmin Bottleneck.}
Fig.~\ref{fig:appendix_softmin} replaces our hard $\min$ bottleneck with a differentiable \texttt{softmin} parameterized by temperature $\tau$.
High temperatures approximate linear mixing (causing collapse), while extremely low temperatures approximate the hard $\min$ but introduce numerical instability.
The hard $\min$ naturally provides the most robust gating mechanism without requiring temperature tuning.

\subsection{Hyperparameter Sensitivity and Baseline Tuning}

\begin{figure}[hbt!]
    \begin{center}
        \centerline{\includegraphics[width=1.0\linewidth]{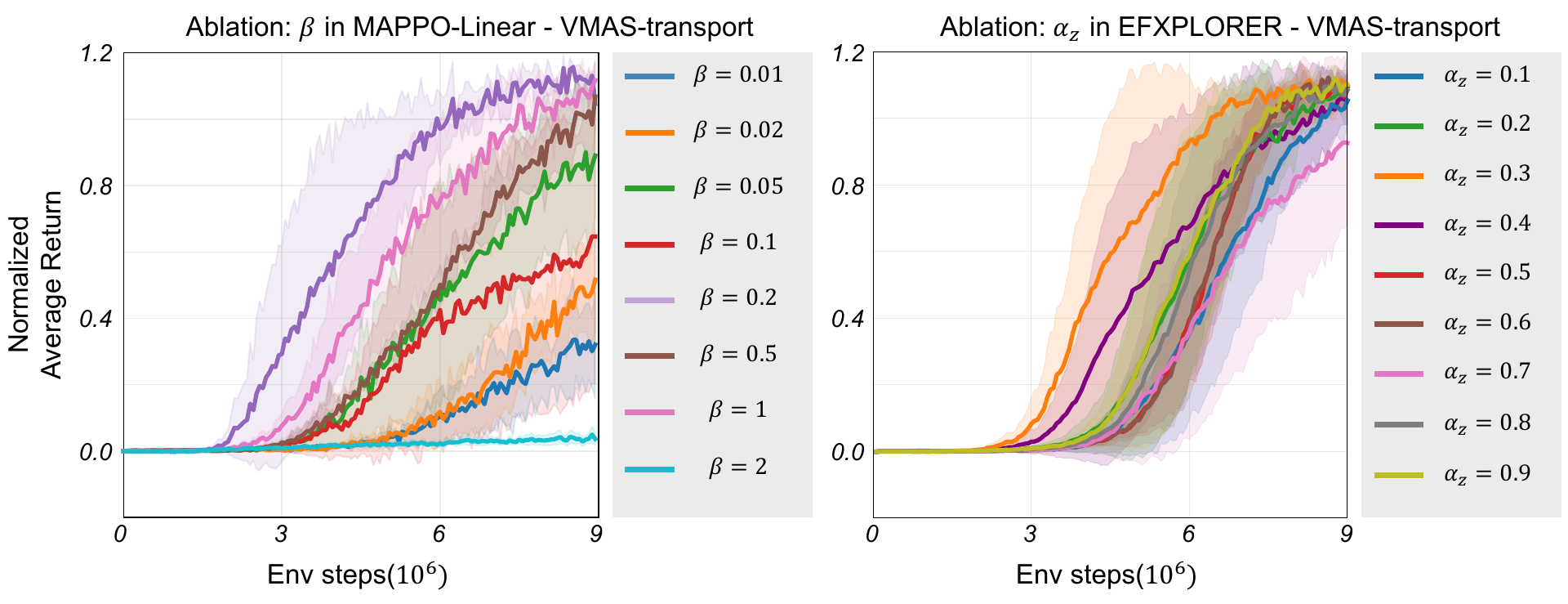}}
    \end{center}
    \vskip -0.3in
    \caption{Sensitivity analysis on \textit{VMAS-transport}. \textsc{EFXPLORER} ($\alpha_z$) robustly converges across a wide range of update rates, whereas MAPPO-Linear ($\beta$) is highly brittle to coefficient changes.}
    \label{fig:appendix_alpha_z}
\end{figure}	
\paragraph{Robustness of Exploration Budgets ($\alpha_z$) vs. Linear Coefficients ($\beta$).}
A core claim of \textsc{EFXPLORER} is its robustness to hyperparameter choices. 
To demonstrate this, we perform a dense grid search over our budget update rate ($\alpha_z$) and the MAPPO-Linear coefficient ($\beta$) on \textit{VMAS-transport} (Fig.~\ref{fig:appendix_alpha_z}). 
Varying $\alpha_z$ only affects exploration \textit{speed}, reliably converging without catastrophic failure. 
Conversely, tuning $\beta$ operates on a razor's edge: slightly high values destroy task performance, while low values yield zero exploration, highlighting the fragility of manual mixing.

\begin{figure}[hbt!]
    \begin{center}
        \centerline{\includegraphics[width=0.72\linewidth]{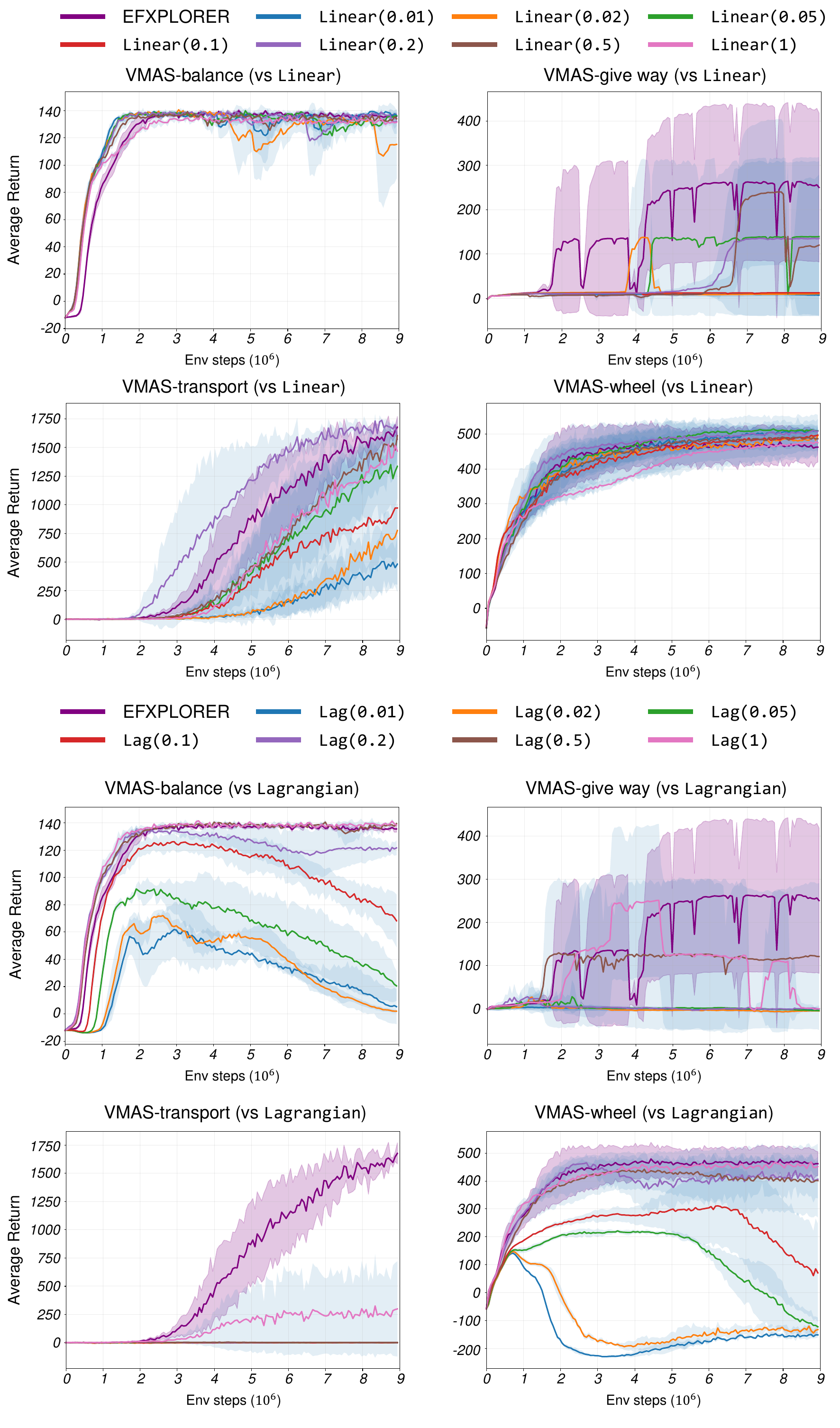}}
    \end{center}
    \vskip -0.3in
    \caption{Comprehensive grid search for baseline hyperparameters ($\beta$ for MAPPO-Linear and $\lambda_0$ for MAPPO-Lagrangian) across four VMAS environments.}
    \label{fig:appendix_linlag_param_search}
\end{figure}

\paragraph{Comprehensive Grid Search for Baselines.}
To ensure fair comparison, Fig.~\ref{fig:appendix_linlag_param_search} reports exhaustive grid searches for MAPPO-Linear ($\beta$) and MAPPO-Lagrangian ($\lambda_0$) across four VMAS environments. 
The results confirm that these standard approaches are highly brittle to hyperparameter choices; unless the coefficients are perfectly tuned, they fundamentally fail to balance task stability with diverse exploration. 
This extreme sensitivity underscores the necessity of our epigraph formulation.

\paragraph{Sensitivity to the Reference Update Rate.}
The running performance reference $\bar{J}_k^{\mathrm{ext}}$ in Eq.~\eqref{eq:outer_feasible_set} determines which rollouts are eligible for budget adaptation.
We therefore examine how its update rate $\alpha_{\mathrm{ref}}$ affects task performance.
The reference is updated as:
\begin{equation}
    \bar{J}_{k+1}^{\mathrm{ext}}
    =
    \begin{cases}
        (1-\alpha_{\mathrm{ref}})\bar{J}_{k}^{\mathrm{ext}}
        + \alpha_{\mathrm{ref}}
        \hat{J}_{k,m^\ast}^{\mathrm{ext}},
        & \mathcal{F}_k \neq \emptyset, \\[2pt]
        \bar{J}_{k}^{\mathrm{ext}},
        & \mathcal{F}_k = \emptyset,
    \end{cases}
    \label{eq:reference_update}
\end{equation}
where $m^\ast$ is the feasible rollout selected by Eq.~\eqref{eq:select_best_rollout}.
To assess sensitivity, we evaluate $\alpha_{\mathrm{ref}}\in\{0.2,0.5,0.8,0.9\}$ on three VMAS tasks, using three seeds per setting.

\begin{table}[h]
    \centering
    \caption{Normalized returns under different reference update rates, using three seeds per setting. The default setting is $\alpha_{\text{ref}}=0.5$.}
    \label{tab:ref_update_rate}
    \begin{tabular}{lcccc}
        \toprule
        \textbf{Environment}    & $\alpha_{\text{ref}}=0.2$ & $\alpha_{\text{ref}}=0.5$ & $\alpha_{\text{ref}}=0.8$ & $\alpha_{\text{ref}}=0.9$ \\
        \midrule
        VMAS-balance                         & $1.03\pm0.011$ & $1.04\pm0.015$   & $1.00\pm0.038$ & $1.03\pm0.017$ \\
        VMAS-transport                         & $1.00\pm0.099$ & $0.98\pm0.130$   & $0.91\pm0.167$ & $0.85\pm0.193$ \\
        VMAS-wheel                         & $1.02\pm0.040$ & $1.03\pm0.029$   & $1.02\pm0.019$ & $1.00\pm0.020$ \\
        \bottomrule
    \end{tabular}
\end{table}

As shown in Table~\ref{tab:ref_update_rate}, performance varies little across the tested rates on VMAS-balance and VMAS-wheel.
On VMAS-transport, larger rates yield lower mean returns and greater variability, suggesting that slower reference updates can be beneficial in this task.
Faster reference updates may make budget adaptation more sensitive to variation in the selected feasible returns.

\paragraph{Hyperparameter and Seed Robustness.}
Comparable best-tuned returns can conceal differences in sensitivity to hyperparameter choices and random seeds.
We therefore conduct a controlled comparison on MPE-tag and VMAS-wheel, where \textsc{EFXPLORER} and tuned MAPPO-Linear achieve comparable final returns.
We evaluate $\beta, \lambda_0\in \{0.01, 0.02, 0.1, 0.5, 1.0\}$ for MAPPO-Linear and MAPPO-Lagrangian, respectively, and $\alpha_z\in \{0.1, 0.3, 0.5, 0.7, 0.9\}$ for \textsc{EFXPLORER}.
Each setting uses five seeds, with all other training configurations held fixed within each task.

We report the final normalized return averaged over settings and seeds.
Steps to $0.8$ denotes the first step at which the averaged learning curve reaches $0.8$.
Seed variance is computed across seeds and averaged over settings, while hyperparameter variance is computed across the seed-averaged final returns of different settings.

\begin{table}[htbp]
    \centering
    \caption{
        Robustness across five hyperparameter settings and five seeds per setting. Steps are reported in millions. N/A indicates that the averaged learning curve does not reach $0.8$.
    }
    \label{tab:hp_seed_robustness}
    \footnotesize
    \setlength{\tabcolsep}{4pt}
    \begin{tabular}{llrrrr}
        \toprule
        \textbf{Environment}
        & \textbf{Method}
        & \shortstack{\textbf{HP-avg. return}}
        & \shortstack{\textbf{Steps to $0.8$ (M)}}
        & \shortstack{\textbf{Seed variance}}
        & \shortstack{\textbf{HP variance}} \\
        \midrule
        \multirow{3}{*}{MPE-tag}
        & MAPPO-Linear
        & 0.99 & 11.70 & 0.0098 & 0.0011 \\
        & MAPPO-Lagrangian
        & 0.69 & N/A & 0.0007 & 0.0031 \\
        & \textsc{EFXPLORER}
        & 1.00 & 11.25 & 0.0069 & 0.0009 \\
        \midrule
        \multirow{3}{*}{VMAS-wheel}
        & MAPPO-Linear
        & 1.00 & 2.61 & 0.0401 & 0.0365 \\
        & MAPPO-Lagrangian
        & 0.42 & 5.50 & 0.1452 & 0.5431 \\
        & \textsc{EFXPLORER}
        & 1.00 & 2.25 & 0.0309 & 0.0178 \\
        \bottomrule
    \end{tabular}
\end{table}

Table~\ref{tab:hp_seed_robustness} shows that \textsc{EFXPLORER} achieves more consistent performance across random seeds and hyperparameter settings than baselines on both tasks.
It also reaches the $0.8$ threshold sooner while maintaining comparable average returns.
MAPPO-Lagrangian has the lowest seed variance on MPE-tag, but its average return is substantially lower.
These results support improved robustness and faster learning as benefits of the epigraph formulation in the tested tasks.

\subsection{Analysis of Exploration and Coordination}\label{app:exploration_coordination}
We examine how exploration budgets adapt across team sizes, whether exploration remains active during temporary reward sacrifice, and how well agents coordinate in SMAX combat scenarios using a task-specific metric.

\paragraph{Exploration-Budget Adaptation across Team Sizes}
Maintaining one exploration budget per agent raises the question of whether budget magnitudes or variation across agents grow excessively with team size.
The budget vector requires $O(N)$ storage and $O(N)$ arithmetic for each elementwise update.
To examine its empirical behavior, we train \textsc{EFXPLORER} on SMAX-\textit{5m\_vs\_6m} ($N=5$) and SMAX-\textit{10m\_vs\_11m} ($N=10$).
At training iteration $k$, we summarize the exploration budgets $\{z_0^{i,k}\}_{i=1}^{N}$ updated by Eq.~\eqref{eq:init_z_ema}.
We report their mean and coefficient of variation (CV) across agents, where CV is the standard deviation divided by the mean.

\begin{table}[h]
    \centering
    \caption{Exploration-budget statistics at three training checkpoints.}
    \label{tab:exp-budget-stat}
    \begin{tabular}{lcccc}
        \toprule
        \textbf{Env. steps}    & Mean budget, $N=5$ & Mean budget, $N=10$ & CV, $N=5$ & CV, $N=10$ \\
        \midrule
        1M & 0.94 & 0.98 & 7.4\% & 8.0\% \\
        10M & 1.92 & 2.01 & 5.0\% & 9.1\% \\
        20M & 2.58 & 1.74 & 5.3\% & 12.5\% \\
        \bottomrule
    \end{tabular}
\end{table}

Table~\ref{tab:exp-budget-stat} shows that per-agent budgets remain in a comparable range across the two team sizes, with CV below $13\%$ at all reported checkpoints.
These results suggest stable budget adaptation across the tested team sizes.
Note that the mean budget for $N=10$ decreases between 10M and 20M steps.
This is consistent with Eq.~\eqref{eq:init_z_ema}, which adapts budgets to the selected feasible rollout without enforcing monotonic growth.

\paragraph{Exploration during Temporary Reward Sacrifice.}
\label{app:temp_sacrifice}

Task-preserving exploration should allow short-term disadvantages when they are compensated by favorable longer-term outcomes.
Our constraint uses reference advantages that account for future returns.
We examine whether budget adaptation keeps intrinsic updates active when one-step and longer-horizon advantage estimates disagree.
Specifically, we use the following event as an empirical proxy
for temporary reward sacrifice:
\begin{equation}
    \mathcal{T}_t
    =
    \left\{
        \hat{A}_{t,1}^{\mathrm{ext}}<0
        \;\land\;
        \hat{A}_{t}^{\mathrm{ext\text{-}GAE}}>0
    \right\},
    \label{eq:temporary_sacrifice_event}
\end{equation}
where $\hat{A}_{t,1}^{\mathrm{ext}}$ is the one-step extrinsic advantage and $\hat{A}_{t}^{\mathrm{ext\text{-}GAE}}$
is the GAE estimate~\cite{schulman2015high}.
This event captures short-term disadvantage relative to the value baseline together with a favorable longer-horizon estimate.

Let $b_{i,t}^{\mathrm{int}}\in\{0,1\}$ indicate whether the intrinsic branch is selected for agent $i$ at timestep $t$.
We report
\begin{equation}
    \rho_{\mathrm{temp}}
    =
    \frac{
        \sum_{i,t}\mathbf{1}[\mathcal{T}_t]
    }{
        \sum_{i,t}
        \mathbf{1}[\hat{A}_{t,1}^{\mathrm{ext}}<0]
    },
    \qquad
    P_{\mathrm{int}\mid\mathrm{temp}}
    =
    \frac{
        \sum_{i,t}
        b_{i,t}^{\mathrm{int}}
        \mathbf{1}[\mathcal{T}_t]
    }{
        \sum_{i,t}\mathbf{1}[\mathcal{T}_t]
    }.
    \label{eq:temporary_sacrifice_metrics}
\end{equation}
Thus, $\rho_{\mathrm{temp}}$ measures how often the longer-horizon estimate becomes favorable among negative one-step-advantage samples.
$P_{\mathrm{int}\mid\mathrm{temp}}$ measures how often the intrinsic update remains active within this subset.
We compare \textsc{EFXPLORER} with an $\alpha_z=0$ ablation on VMAS-transport and VMAS-wheel.
The ablation fixes the exploration budgets at their initial values, allowing us to examine the contribution of budget adaptation.

\begin{table}[htbp]
    \centering
    \caption{
        Exploration during temporary reward sacrifice. $\rho_{\mathrm{temp}}$ and $P_{\mathrm{int}\mid\mathrm{temp}}$ are defined in Eq.~\eqref{eq:temporary_sacrifice_metrics}. Steps are reported in millions of environment interactions. The $\alpha_z=0$ ablation disables budget adaptation.
    }
    \label{tab:temporary_sacrifice}
    \footnotesize
    \setlength{\tabcolsep}{4pt}
    \begin{tabular}{lclrrr}
        \toprule
        \textbf{Environment}
        & \shortstack{\textbf{Steps (M)}}
        & \textbf{Variant}
        & $\rho_{\mathrm{temp}}$
        & $P_{\mathrm{int}\mid\mathrm{temp}}$
        & \shortstack{\textbf{Normalized return}} \\
        \midrule
        \multirow{6}{*}{VMAS-transport}
        & \multirow{2}{*}{3}
        & \textsc{EFXPLORER} & 0.315 & 0.288 & 0.018 \\
        & & $\alpha_z=0$ & 0.250 & 0.025 & 0.004 \\
        \cmidrule(lr){2-6}
        & \multirow{2}{*}{6}
        & \textsc{EFXPLORER} & 0.563 & 0.801 & 0.762 \\
        & & $\alpha_z=0$ & 0.295 & 0.207 & 0.164 \\
        \cmidrule(lr){2-6}
        & \multirow{2}{*}{9}
        & \textsc{EFXPLORER} & 0.392 & 0.764 & 1.065 \\
        & & $\alpha_z=0$ & 0.409 & 0.445 & 0.797 \\
        \midrule
        \multirow{6}{*}{VMAS-wheel}
        & \multirow{2}{*}{3}
        & \textsc{EFXPLORER} & 0.575 & 0.374 & 0.914 \\
        & & $\alpha_z=0$ & 0.513 & 0.037 & 0.843 \\
        \cmidrule(lr){2-6}
        & \multirow{2}{*}{6}
        & \textsc{EFXPLORER} & 0.470 & 0.581 & 0.982 \\
        & & $\alpha_z=0$ & 0.466 & 0.007 & 0.946 \\
        \cmidrule(lr){2-6}
        & \multirow{2}{*}{9}
        & \textsc{EFXPLORER} & 0.464 & 0.531 & 1.008 \\
        & & $\alpha_z=0$ & 0.457 & 0.006 & 0.964 \\
        \bottomrule
    \end{tabular}
\end{table}

Table~\ref{tab:temporary_sacrifice} shows that temporary-sacrifice samples occur in both variants.
Within these samples, \textsc{EFXPLORER} selects the intrinsic branch more frequently than the fixed-budget ablation.
It also achieves higher returns at every reported checkpoint.
These results suggest that the budget adaptation helps retain intrinsic updates when the one-step task advantage is negative but the longer-horizon estimate is favorable.

To examine how often each branch contributes to the update, we track the intrinsic indicator $b_{i,t}^\text{int}$.
The intrinsic branch is selected when $b_{i,t}^\text{int}=1$, and the extrinsic branch when $b_{i,t}^\text{int}=0$.
The latter accounts for at least $30\%$ of samples at every reported checkpoint.
Together with the conditional selection rate in Table~\ref{tab:temporary_sacrifice}, this shows that the intrinsic updates remain active in temporary-sacrifice cases while the extrinsic branch continues to contribute throughout training.

\paragraph{Quantifying Coordination in SMAX.}
\label{app:quantify_coordination}

Returns and success rates often provide limited information about how agents coordinate during exploration. 
To complement these outcome metrics, we introduce a task-specific coordination proxy on SMAX-\textit{5m\_vs\_6m}.
In this scenario, the allied Marines are numerically inferior, making collective engagement and kiting particularly relevant.
We compare \textsc{EFXPLORER} with MAPPO-Linear using $\beta=0.1$ and $\beta=1.0$.

To quantify local numerical balance during combat, let $A_t$ be the number of attacking allies and $E_t$ the number of enemies within allied attack range.
Motivated by the importance of relative force size in Lanchester-style models of StarCraft combat~\cite{stanescu2015using}, we define a local engagement score:
\begin{equation}
    \mathcal{T}_{\mathrm{eng}}
    =
    \{t:A_t>0,\ E_t>0\},
    \qquad
    C_{\mathrm{eng}}
    =
    \frac{1}{|\mathcal{T}_{\mathrm{eng}}|}
    \sum_{t\in\mathcal{T}_{\mathrm{eng}}}
    \mathbf{1}[A_t\geq E_t].
    \label{eq:engagement_coordination}
\end{equation}
A high $C_{\mathrm{eng}}$ indicates that allied attacks occur more frequently when attacking allies are at least as numerous as enemies within range.
This is a proxy for local engagement structure rather than a direct measure of shared target selection.
We report $C_{\mathrm{eng}}$ alongside normalized return and episode length.

\begin{table}[htbp]
    \centering
    \caption{
        Local engagement structure and task performance on SMAX-5m\_vs\_6m. $C_{\mathrm{eng}}$ is defined in Eq.~\eqref{eq:engagement_coordination}. Episode length is measured in environment timesteps.
    }
    \label{tab:coordination_smax}
    \footnotesize
    \setlength{\tabcolsep}{5pt}
    \begin{tabular}{llrrr}
        \toprule
        \textbf{Env.\ steps}
        & \textbf{Method}
        & $C_{\mathrm{eng}}\,\uparrow$
        & \shortstack{\textbf{Normalized return}}
        & \shortstack{\textbf{Episode length}} \\
        \midrule
        \multirow{3}{*}{5M}
        & \textsc{EFXPLORER}
        & 75\% & 0.23 & 40.0 \\
        & MAPPO-Linear ($\beta=0.1$)
        & 33\% & 0.20 & 14.5 \\
        & MAPPO-Linear ($\beta=1.0$)
        & 28\% & 0.23 & 13.9 \\
        \midrule
        \multirow{3}{*}{10M}
        & \textsc{EFXPLORER}
        & 67\% & 0.34 & 45.3 \\
        & MAPPO-Linear ($\beta=0.1$)
        & 31\% & 0.28 & 14.7 \\
        & MAPPO-Linear ($\beta=1.0$)
        & 25\% & 0.28 & 14.1 \\
        \midrule
        \multirow{3}{*}{15M}
        & \textsc{EFXPLORER}
        & 80\% & 0.61 & 78.5 \\
        & MAPPO-Linear ($\beta=0.1$)
        & 27\% & 0.30 & 14.4 \\
        & MAPPO-Linear ($\beta=1.0$)
        & 27\% & 0.31 & 14.1 \\
        \midrule
        \multirow{3}{*}{20M}
        & \textsc{EFXPLORER}
        & 80\% & 0.80 & 65.0 \\
        & MAPPO-Linear ($\beta=0.1$)
        & 30\% & 0.31 & 14.5 \\
        & MAPPO-Linear ($\beta=1.0$)
        & 22\% & 0.30 & 13.9 \\
        \bottomrule
    \end{tabular}
\end{table}

As shown in Table~\ref{tab:coordination_smax}, \textsc{EFXPLORER} maintains higher engagement scores and longer episodes than both linear-mixing settings at every checkpoint.
By 20M steps, its normalized return reaches $0.80$ with a $60\%$ win rate, while baseline returns remain near $0.30$.

The observed rollouts help explain these differences.
With $\beta=0.1$, MAPPO-Linear shows limited discovery of effective kiting behavior.
With $\beta=1.0$, agents engage separately or split their attacks, leading to fragmented engagements and early elimination.
In contrast, \textsc{EFXPLORER} forms a compact group, draws smaller subsets of enemies forward, and attacks under local numerical advantage.
Together, the quantitative and qualitative results support exploration that discovers new tactics while maintaining coordinated engagement in this scenario.
%%%%%%%%%%%%%%%%%%%%%%%%%%%%%%%%%%%%%%%%%%%%%%%%%%%%%%%%%%%%%%%%%%%%%%%%%%%%%%%
%%%%%%%%%%%%%%%%%%%%%%%%%%%%%%%%%%%%%%%%%%%%%%%%%%%%%%%%%%%%%%%%%%%%%%%%%%%%%%%

%\newpage
%\input{checklist}

% Disable again so the checklist does not appear in the ToC
\renewcommand{\addcontentsline}[3]{}

\end{document}